\documentclass[conference]{IEEEtran}
\usepackage{times}
\usepackage[numbers]{natbib}
\usepackage{multicol}
\usepackage{bm}
\usepackage{amsmath}
\usepackage{booktabs}
\usepackage{graphicx}
\usepackage{float}
\usepackage{algorithm2e}
\usepackage[usenames,dvipsnames,table,xcdraw]{xcolor}
\usepackage[resetlabels,labeled]{multibib}
\newcites{A}{Appendix References}

\usepackage[font=footnotesize,labelfont=bf]{caption}
\usepackage{makecell}

\usepackage{hyperref}
\hypersetup{
    colorlinks=true,
    linkcolor=MidnightBlue,
    filecolor=MidnightBlue,      
    urlcolor=MidnightBlue,
    citecolor=MidnightBlue,
}

\begin{document}

\title{SpringGrasp: An optimization pipeline for robust and compliant dexterous pre-grasp synthesis}
\title{SpringGrasp: Robust and compliant dexterous pre-grasp synthesis under shape uncertainty}
\title{SpringGrasp: Synthesizing Compliant, Dexterous Grasps under Shape Uncertainty}

\author{\authorblockN{Sirui Chen,
Jeannette Bohg,
C. Karen Liu}
\authorblockA{Department of Computer Science, Stanford University
\\ \{ericcsr, bohg, karenliu\}@cs.stanford.edu}}



\maketitle

\begin{abstract}

Generating stable and robust grasps on arbitrary objects is critical for dexterous robotic hands, marking a significant step towards advanced dexterous manipulation. Previous studies have mostly focused on improving differentiable grasping metrics with the assumption of precisely known object geometry. However, shape uncertainty is ubiquitous due to noisy and partial shape observations, which introduce challenges in grasp planning. 
We propose, SpringGrasp planner, a planner that considers uncertain observations of the object surface for synthesizing compliant dexterous grasps. A compliant dexterous grasp could minimize the effect of unexpected contact with the object, leading to more stable grasp with shape-uncertain objects. We introduce an analytical and differentiable metric, SpringGrasp metric, that evaluates the dynamic behavior of the entire compliant grasping process. Planning with SpringGrasp planner, our method achieves a grasp success rate of 89\% from two viewpoints and 84\%  from a single viewpoints in experiment with a real robot on 14 common objects. Compared with a force-closure based planner, our method achieves at least \textcolor{black}{18\%} higher grasp success rate.

\end{abstract}

\IEEEpeerreviewmaketitle

\section{Introduction}
Grasping with multi-fingered dexterous hands is an essential manipulation capability that enables a robot to interact with everyday objects - especially those with irregular shapes, \textcolor{black}{such as different fruits and bottles.} 
Previous work addresses the dexterous grasp planning problem mainly from two perspectives: data-driven grasp generation~\cite{deepgraspsurvey, datagraspsurvey} and optimization based grasp generation using differentiable grasp criteria such as force closure~\cite{diffwc, wubilevel} or min-weight~\cite{frogger}.Most optimization based methods aims to obtain a set of stable contact points on the object surface for each fingertip, which requires a detailed object model to be available.  However, perception in the real world usually relies on noisy sensors with limited viewpoints, resulting in uncertainty about the object shape that poses significant challenges to previous works. Data-driven method can work with partial and noisy observation\cite{datagraspsurvey,deepgraspsurvey}, but most of them focus on simplified problem of choosing a few fixed grasp candidates without optimizing for object and grasp stability.
To grasp objects from noisy and partial observations in the real world, \textcolor{black}{we argue that} every component in the grasping pipeline should consider the impact of shape uncertainty. The object perception model needs to quantify shape uncertainty from the sensor data \cite{gpis, psdf, ppcd}, the grasp planner needs to leverage uncertain object information to make optimal plans \cite{mostsimilar, gp_gpis_opt}, and finally the grasp controller also needs to be compliant to tolerate unexpected contact between fingertips and the object due to shape uncertainty. An optimization based method that can efficiently optimize compliant pregrasps while considering object shape uncertainty is essential for achieving such tasks.

\begin{figure}[t]
  \centering
  \includegraphics[width=0.9\linewidth]{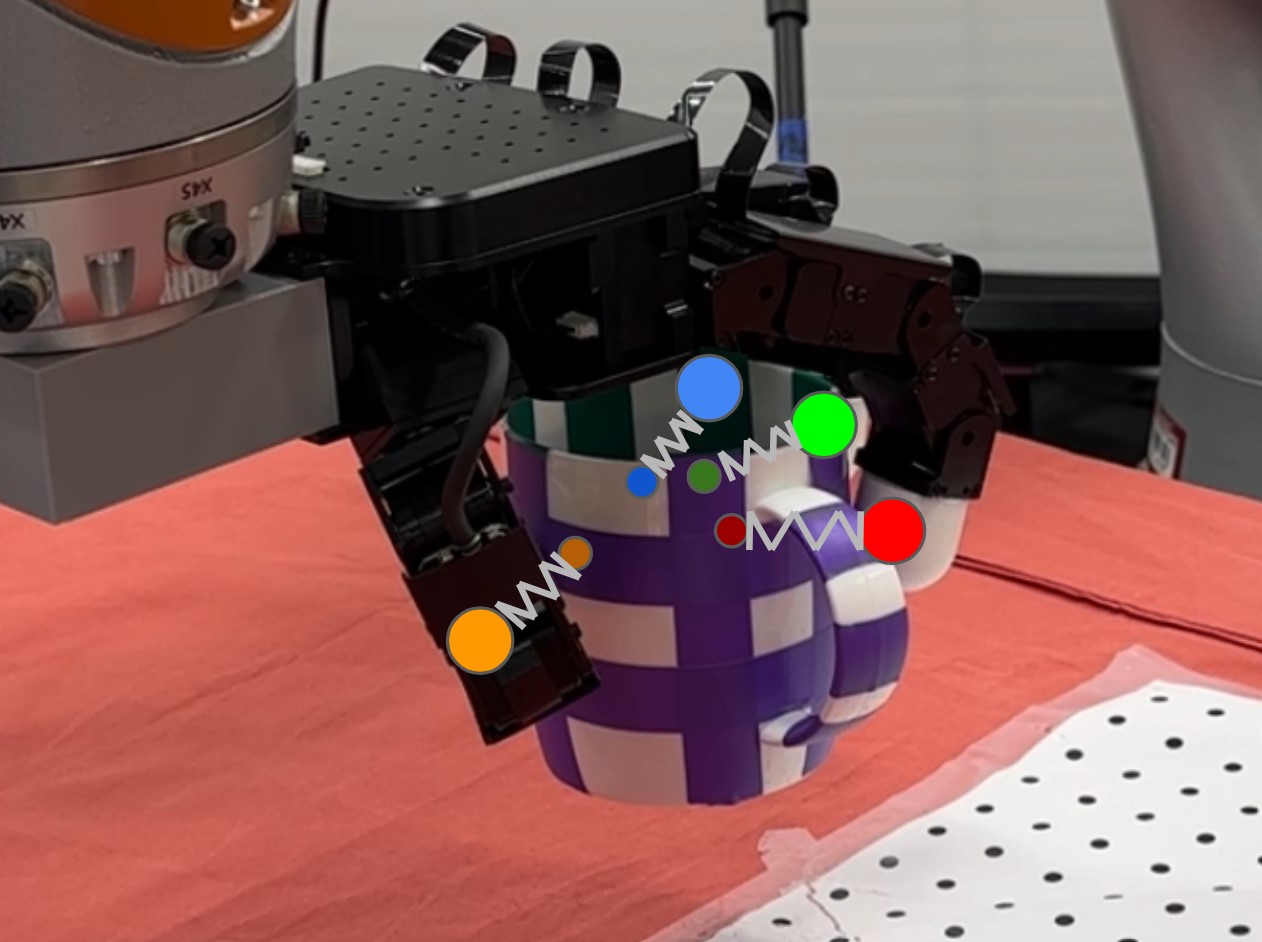}
  \caption{\textcolor{black}{Compliant grasp as virtual spring attached from finger to object}}
  \label{fig:teaser}
\end{figure}

We propose SpringGrasp, a planner that considers uncertain observations of the object surface for synthesizing compliant dexterous grasps. A complaint grasp consists of an optimized pregrasp hand pose and optimized per-finger impedance controls.  Unlike the conventional definition of a grasp, we view a compliant grasp as a dynamic process starting from the pregrasp pose, grasping the object according to the per-finger impedance controls, and moving the object along with the finger motion until a stable equilibrium is reached. As shown in Fig.\ref{fig:teaser}, a compliant grasp in contact can be modelled as virtual springs connected to the object. Critical to the optimization of SpringGrasp, we introduce an analytical and differentiable metric to evaluate whether the compliant grasping process can lead to a force-closure grasp at equilibrium. In contrast to existing methods that solve for contact locations, our optimized compliant grasp allows for direct control over force magnitudes, force directions and impedance parameters. As such, the unexpected contacts with the object can be minimized, leading to more stable manipulation with uncertain object shapes.

We evaluate our method on a real robotic platform consisting of an Allegro hand and RGB-D cameras. We demonstrate that our method can grasp objects from noisy and partial observations with at least 18\% higher success rate compared to baselines. Our method can even achieve 84\% grasp success rate with a single depth camera input thanks to our optimized compliant grasp. 
We will release our grasp planner in a python package at \url{https://github.com/Stanford-TML/SpringGrasp_release}.

\section{Related Work}
\subsection{Optimization based methods for dexterous grasp}

Synthesizing precision grasps by solving an optimization problem has been extensively studied over the years. Most research integrates an analytical grasp metric into the objective function or constraints of the optimization problem~\cite{metricsurvey}, utilizing either gradient-based or sampling-based optimizers. Recent studies have focused on optimizing robust and kinematically reachable contact points between the hand and object, using either an explicit mesh model of the object \cite{wubilevel, frogger} or an implicit signed distance function (SDF) \cite{dexgraspnet, diffwc}. However both, a detailed mesh or SDF are challenging to obtain in the real world due to noisy sensors and partial observability. 

The design of the grasp metric is crucial, as it determines both the quality of the synthesized grasp and the optimization runtime. Recent works \cite{diffwc, frogger} have demonstrated that gradient-based optimization can significantly accelerate the grasp optimization process. Therefore, our comparison focuses on differentiable grasp metrics, among which the most widely used is the differentiable force closure criterion~\cite{forceclosure, wubilevel, diffwc}. Given a set of contact locations ${\bm{p}_i}$ and normals ${\bm{n}_i}$, this criterion assesses whether there exists a set of nonzero contact forces ${\bm{F}_i}$ at each fingertip that can achieve force and torque equilibrium $\sum_i F_i = 0, \sum_i \bm{r}\times F_i = 0$ within the Coulomb friction cone $\bm{F}_i\cdot\bm{n}_i>\frac{1}{\sqrt{1+\mu}}||\bm{F}_i||$. An alternative metric, the Min-weight metric~\cite{frogger}, also measures the robustness of a grasp under force perturbations and speeds up convergence during optimization. However, evaluating force closure involves solving a convex optimization problem, which is not analytically solvable. Therefore, optimizing a grasp with the force closure criterion and under kinematics constraints often necessitates bi-level optimization, which is typically slow to evaluate and challenging to optimize using off-the-shelf gradient-based optimizers \cite{frogger,diffwc, bilevelsurvey}. DexGraspNet~\cite{dexgraspnet} employs an analytical criterion to approximate force closure, but this does not guarantee force closure of the solution. As an alternative to bi-level optimization, there are also many works that sample initial pregrasps using various heuristics and then either optimize a final grasp from this starting point or do rejection sampling based on the force closure criterion. We refer to~\cite{datagraspsurvey} for more details. \textcolor{black}{Our method uses a novel compliant grasp formulation that models the grasp as a dynamic process and a novel differentiable grasp metric that measures contact feasibility during the process and force closure at the equilibrium state.}

\subsection{Dexterous grasping under shape uncertainty}
In many scenarios, a robot does not have access to accurate object models and must rely on perceiving the object through its depth sensors. The resulting perceived object model often contains shape uncertainties due to sensor noise and partial observability. As demonstrated for example in \cite{mostsimilar,gpis,gpis_tactileexploration}, grasping a shape-uncertain object with a dexterous hand requires a system that considers shape uncertainty arising from perception.
\subsubsection{Grasping shape uncertain objects with compliance}
Compliance plays a crucial role in enabling robots to robustly grasp and manipulate objects with varying geometries \cite{surprising,PisaSoftHand,EnvConstraints}. Numerous studies have demonstrated this capability using passive compliant multi-fingered hands \cite{rbohand,pisaiit,oceanonehand,PisaSoftHand}, where compliance is derived from the mechanical design. However, passive compliance often leads to a reduction in the number of actuated degrees of freedom and poses challenges in accurately modeling kinematics. Both of these issues make it challenging to synthesize and execute precision grasps. 
In contrast, active compliant multi-fingered hands \cite{biorealistic, fingertipcompliance, mostsimilar, external_contact} address these limitations by simulating compliance through control of motorised joints. Many previous works have focused on power grasps with passive compliance \cite{rbohand, surprising, pisaiit,PisaSoftHand,EnvConstraints}. However, in this work we are interested in compliant precision grasps such as considered in for example \cite{mostsimilar,fingertipcompliance,doi:10.1177/0278364911416526}. Among these works \cite{mostsimilar} adapt joint level compliance based on surface variance to achieve a precision grasp while \cite{doi:10.1177/0278364911416526, fingertipcompliance} focus on how to control the object compliantly after a precision grasp. Our work focus on generating grasps using Cartesian space impedance control with optimized gains. At the same time, impedance control with different gains optimized for different tasks has been shown effective in contact rich manipulation tasks and robot locomotion \cite{variableimpedance, taskimpedance}, while \cite{variableimpedance} also consider changing the impedance through time, Our method automatically determined appropriate controller gains per grasp and object.  Instead of picking impedance controller gain manually as in \cite{taskimpedance} or choosing by reinforcement learning in \cite{variableimpedance}, our method optimize controller gains based on each object's uncertainty and geometry.

\subsubsection{Modelling shape uncertain objects}
Objects with shape uncertainty have often been represented as probabilistic occupancy maps in environmental mapping problems \cite{proboccupancy, gridslam}. The study in \cite{sample_repr} develops a probabilistic camera model and utilizes a set of samples to represent objects with shape uncertainty. However, both grid maps and sample-based representations are memory-intensive, posing challenges in detailed 3D object representation. To overcome this, recent methods have employed probabilistic implicit functions, such as Gaussian Process Implicit Surfaces (GPIS) \cite{gpis, gpis_tactile, gpis_tactileexploration, gp_gpis_opt} or Probabilistic Signed Distance Functions (pSDF) \cite{psdf}. Studies like \cite{mostsimilar, gpis_tactile} use GPIS with thin plate kernel functions to smoothly extrapolate the unobserved surface. Rather than directly representing objects with implicit functions, \cite{gp_filter} employs Gaussian processes to filter out noisy points, modeling the object using only points with low uncertainty. More recently, various shape completion models have utilized deep neural networks to complete object shapes with greater detail \cite{shapecompletion, pfnet, pvd, diffusionsdf}. Given the potential ambiguity in object shapes from limited views, \cite{pvd, diffusionsdf} employ generative models to provide multiple plausible guesses based on point clouds from a single view. \textcolor{black}{Similar to \cite{mostsimilar}, our method uses GPIS to model uncertain object shape from a partially observed point cloud.}

\subsubsection{Dexterous grasp planning under surface uncertainty}
Using above mentioned object modelling, some works plan grasps based on shape completed objects \cite{graspshapecompletion, graspdropout}. Other works \cite{mostsimilar, gp_gpis_opt, pong} plan grasps with probabilistic object model and consider uncertainty explicitly. Such work typically necessitates an object model that can quantify uncertainty at various positions, such as GPIS \cite{mostsimilar}. Research by \cite{gp_gpis_opt, mostsimilar} aims to minimize the impact of uncertainty in grasp planning. Both studies penalize making contact in regions of high variance. Additionally, \cite{mostsimilar} scales down the force applied at each fingertip in proportion to the uncertainty at the contact point. Under shape uncertainty, achieving precise contact at a planned location is challenging. Some studies refine the contact location and force after the fingertip makes contact with the object \cite{mostsimilar, fingeradapt}. Others iteratively explore the object's shape and re-plan the grasping pose using tactile probing \cite{gpis_tactileexploration, gpis_tactile}. However, these approaches require additional information from tactile sensors, which are expensive and add fragility to dexterous multi-finger hands. Employing a pre-grasp hand pose before actual grasping can prevent accidental contact with the object and enhance the stability of the subsequent grasp \cite{dmppregrasp, preshape, multipregrasp}. \cite{multipregrasp} demonstrates that a carefully chosen pre-grasp can even facilitate grasping multiple loosely piled objects. While most studies rely on manually defined pre-grasp poses without specific planning \cite{preshape, pregraspopposition}, research like \cite{multipregrasp, dmppregrasp} plans pre-grasps using a force closure-based grasp planner with an inflated object model. In our work, we approximate object shape using GPIS based on point clouds from either one or multiple views. Different from previous works, we propose a new grasp metric SpringGrasp metric that can optimize pregrasp when considering uncertainty and adapt impedance gains according to object uncertainty and geometry.

\section{Method}
\begin{figure*}[t]
  \centering
  \includegraphics[width=\linewidth]{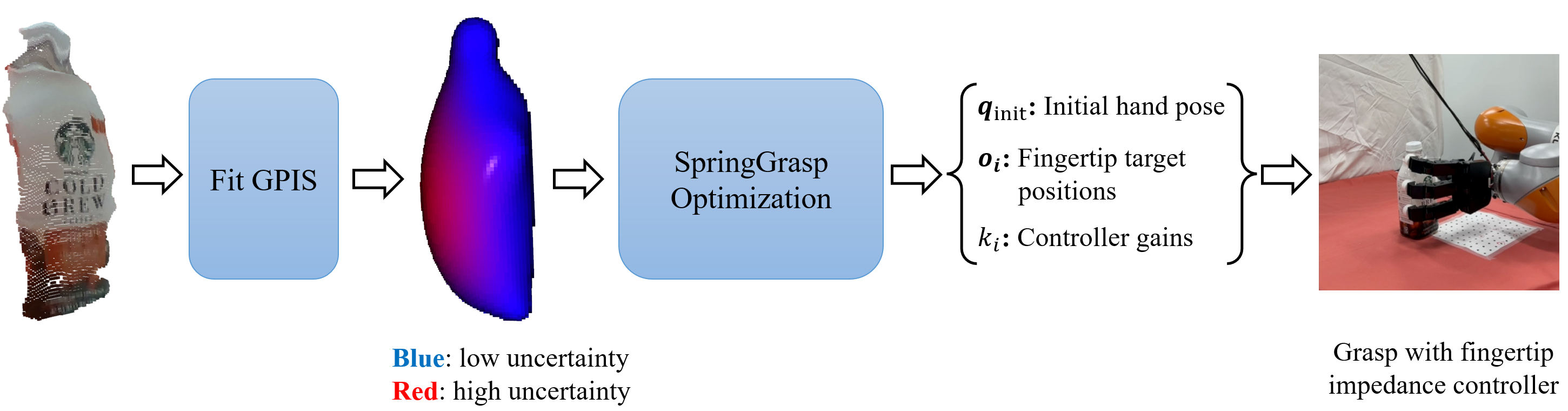}
  \caption{\textcolor{black}{Process for generating grasps from a partial point cloud.}}
  \label{fig:model}
\end{figure*}

We introduce an optimization-based grasp planner, SpringGrasp planner,capable of grasping objects with shape uncertainty. Our planner is designed to directly interface with objects represented as point clouds, either sourced from depth cameras or generated by a shape completion network. We formulate an optimization that solves for a compliant grasp which involves a pregrasp pose and per-finger impedance controls that take into account the potential motion of the object during the process of grasping. Critical to the optimization, we introduce a novel analytical and differentiable metric, SpringGrasp metric,to evaluate whether the compliant grasp can reach a stable equilibrium. In addition, our optimization takes into account the uncertainty of the object surface when planning the compliant grasp. Fig.~\ref{fig:model} illustrate the process of planning and executing a compliant pre-grasp using SpringGrasp planner.


\subsection{Compliant grasp formulation}
\label{subsec:formulation}
The compliant grasp is defined as $\mathcal{G} = \{\bm{p}_i(t_0), \bm{o}_i, k_i\}, i\in \{1,2,...,m\}, \textcolor{black}{m\ge 3}$, which include a set of $m$ fingertip initial contact locations on object surface $\bm{p}_i(t_0)\in\mathcal{R}^3$ and target locations $\bm{o}_i\in \mathcal{R}^3$ both in the world frame and a set of controller gains $k_i\in \mathcal{R}$. 

\begin{figure}[t]
  \centering
  \includegraphics[width=0.8\linewidth]{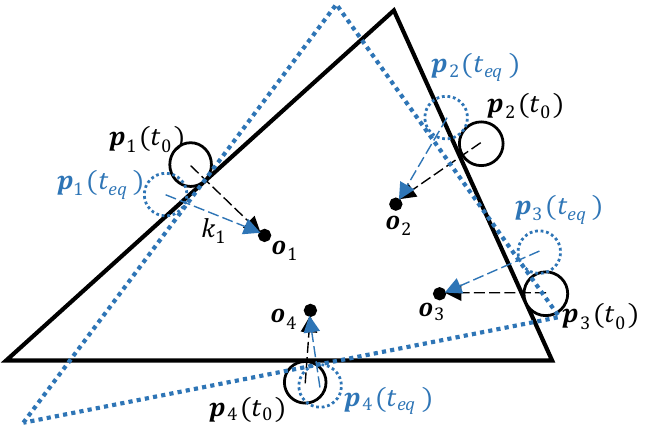}
  \caption{A compliant grasp $\mathcal{G}$ on a 2D triangle at $t_0$ and at $t_\text{eq}$. The black solid line shows initial fingertip positions and the object pose, the blue dashed line shows fingertip positions and the object pose at equilibrium. $\bm{p}_i$ is the fingertip contact position and $\bm{o}_i$ is the target position.}
  \label{fig:compliantgrasp}
\end{figure}

For each finger, starting from the initial state contact position $\bm{p}_i(t_0)$, we control the position of the fingertip using a Cartesian-space impedance controller moving toward $\bm{o}_i$ with a gain $k_i$ and damping coefficient $2\sqrt{k_i}$. We do not require that the contacts of initial state at $\bm{p}_i(t_0)$ to form a force-closure grasp. Instead, we expect the pose of the object to change over time in a dynamic process until an equilibrium is reached at $t_\text{eq}$. We assume that fingertips will move with the object without slippage during this dynamic process from $t_0$ to $t_{eq}$. Fig. \ref{fig:compliantgrasp} shows a 2D example of our compliant grasping process of a triangle.

The effect of each fingertip on the object can be simulated as a virtual spring-damper system attached to the object between $\bm{p}_i(t)$ and $\bm{o}_i$. As such, we model the compliant grasping process as a simultaneous motion of the object and springs toward the equilibrium state through time. We assume that the contact points in the object frame $\bm{p}^o_i = \bm{R}^T_0(\bm{p}_i(t_0) - \bm{t}_0)$ are fixed over time (i.e. no sliding), where $\bm{t}_0, \bm{R}^T_0$ are the known translation and rotation from the world to the object frame at $t_0$. 

Since we are interested to know whether a force-closure grasp can be formed at $t_{eq}$, we need to obtain the fingertip positions $\bm{p}_i(t_{eq})$. A naive way to obtain $\bm{p}_i(t_{eq})$ is to forward simulate the grasping process from $t_0$ to $t_\text{eq}$. However, introducing a numerical forward simulation process to the optimization adds significant computational cost and numerical instability due to gradient back-propagation\cite{pods,diffrl}. Our key insight is that we can express the fingertip positions at equilibrium by the transformation of the object $\bm{R}_\text{eq}, \bm{t}_\text{eq}$:
\begin{equation}
\label{eqn:fingertips_equil}
    \bm{p}_i(t_{eq}) = \bm{R}_{eq}\bm{R}^T_0\bm{p}_i(t_0) + \bm{t}_{eq} - \bm{R}_{eq}\bm{R}^T_0\bm{t}_{0},
\end{equation}
and formulate a Wahba's problem\cite{wahba} which solves for the optimal rigid transformation of the object to minimize total kinetic energy stored in each virtual spring with gain $k_i$:
\begin{equation}
\label{eqn:rigid_transformation_opt}
    \bm{t}_\text{eq}, \bm{R}_\text{eq} = \underset{\bm{t}, \bm{R}}{\arg\min}\ \sum_i^m\frac{1}{2}k_i||\bm{R}\bm{R}^T_0(\bm{p}_i(t_0) - \bm{t}_0) + \bm{t} - \bm{o}_i||_2^2.
\end{equation}

Wahba's problem can be solved analytically with a modified Kabsch algorithm~\cite{kabsch, wahbaanalytical}. Alg. \ref{alg:optimaltf} shows detailed implementation of the analytical solver. Combining Equation \ref{eqn:fingertips_equil} and Equation \ref{eqn:rigid_transformation_opt}, we can now express $\bm{p}_i(t_{eq})$ as a function of $\bm{p}_i(t_0)$, $\bm{o}_i$, and $k_i$.

\RestyleAlgo{ruled}
\SetKwComment{Comment}{/* }{ */}
\begin{algorithm}
\caption{Modified Kabsch algorithm}\label{alg:optimaltf}
\(\textbf{Input:}\;\bm{p}_i(t_0),\bm{o}_i,k_i,i\in\{1,2,...,m\}\) \\
\(\textbf{Output:}\;\bm{t}_{\text{eq}}, \bm{R}_{\text{eq}}\) \\
\(\bm{c}_p \gets \frac{1}{m}\sum_i \bm{p}_i(t_0)\)\\
\(\bm{c}_o \gets \frac{1}{m}\sum_i \bm{o}_i\)\\
\(\bm{P} \gets \text{Stack}(\{\bm{p}_i(t_0) - \bm{c}_p\})\) \Comment*[r]{mx3 matrix}
\(\bm{O} \gets \text{Stack}(\{\bm{o}_i - \bm{c}_o\})\) \Comment*[r]{mx3 matrix}
\(\bm{K} \gets \text{Stack}(\{k_i\})\) \Comment*[r]{ 1xm matrix}
\(\bm{H} \gets (\bm{K}\bm{P})^T(\bm{K}\bm{O})\)\\
\(\bm{U}, \bm{V} \gets SVD(\bm{H})\)\\
\(\bm{R} \gets \bm{V}\bm{U}^T\)\\
\If{\(\text{det}(\bm{R}) < 0\)}{
    \(V[:,-1] \gets -V[:,-1]\) \Comment*[r]{prevent mirroring}
}
\(\bm{R}_{\text{eq}} \gets \bm{V}\bm{U}^T\)\\
\(\bm{t}_{\text{eq}} \gets \frac{1}{\sum_i k_i}\sum_i k_i(\bm{o}_i - \bm{R}_{\text{eq}}\bm{p}_i(t_0))\)\\
\end{algorithm}

\subsection{SpringGrasp metric}
With an analytical expression of $\bm{p}_i(t_{eq})$, we can now define a metric that determines whether the fingertips at equilibrium can form a force-closure grasp. For each fingertip, we define the contact margin $\epsilon_i(t)$ as the angular difference between the direction of force $\bm{f}_i(t)$ and the surface normal $\bm{n}_i(t)$ at the contact in the world frame:
\begin{equation}
\epsilon_i(t) = -\frac{\bm{f}_i(t)}{||\bm{f}_i(t)||^2_2}\cdot\bm{n}_i(t) - \frac{1}{\sqrt{1+\mu^2}}, i\in\{1,2,...,m\},
\end{equation}
where the contact force, expressed in the world frame, is modelled as the virtual spring force:
\begin{equation}
\bm{f}_i(t) = k_i(\bm{o}_i - \bm{p}_i(t)) - 2\sqrt{k_i}\dot{\bm{p}}_i(t), i\in \{1,2,...,m\}.
\end{equation}

If the contact margins on all fingertips are non-negative during the grasping process, that is, $\epsilon_i(t)\ge 0, i\in{1,2,...,m}, t\in[t_0,t_\text{eq}]$, the object can be successfully grasped. However, when the difference between $\bm{R}_\text{eq}$ and $\bm{R}_0$ (i.e. the object rotation during the compliant grasping process) is small, enforcing the non-negative contact margins at the initial and equilibrium states, $\epsilon_i(t_0) \geq 0$ and $\epsilon_i(t_{eq}) \geq 0$, guarantees that all states in between also have non-negative contact margins (See Appendix \ref{sec:apped_proof} for detailed analysis).

\subsection{Optimizing compliant grasp with known object surface}
With the SpringGrasp metric, we can formulate an optimization to solve for a compliant grasp. We first describe the formulation for grasping an object with known surface in this subsection, and extend it to grasping an object with uncertain surface in the next subsection. We define our SpringGrasp metric energy as
\begin{equation}
\label{eqn:Esp}
E_\text{sp} = -\sum_i^m\log(\epsilon_i(t_0)+1)-\sum_i^m\log(\epsilon_i(t_\text{eq})+1)
\end{equation}
Let the surface of the object be a Signed Distance Function (SDF) $d(\bm{x})$, which maps a 3D point $\bm{x}$ to the signed distance between $\bm{x}$ and the object surface. We solve for a compliant grasp $\mathcal{G} = \{\bm{p}_i(t_0), \bm{o}_i, k_i\}, i\in \{1,2,...,m\}$ that minimizes the distance between fingertips to the object surface, $E_\text{dist} = |d(\bm{p}_i(t_0))|$, while optimizing the SpringGrasp metric energy:
\begin{equation}
\label{eqn:springgrasp_only_opt}
    \underset{\{\bm{p}(t_0)\},\{\bm{o}\},\{k\}}{\arg\min}\; E_\text{sp} + E_\text{dist},
\end{equation}
where $\{x\}$ is a shorthand for the set $x_i, i \in\{1,2,\cdots,m\}$.

The logarithm formulation in $E_\text{sp}$ (Equation \ref{eqn:Esp}) encourages the optimizer to focus on fingertips with negative margins instead of improving fingertips that already have positive margins. In rare cases when margin is below -1, Appendix. \ref{sec:clip_esp} shows how we handle them with clipping and adding auxiliary energy terms.

\subsection{Pregrasp with uncertain object surface}
\label{subsec:gpis}
The optimization described in Equation \ref{eqn:springgrasp_only_opt} assumes that the object surface is precisely known, which is often not true in the real-world scenarios. 
To model objects with shape uncertainty, we use the widely adopted {\em Gaussian Process Implicit Surface\/} (GPIS) to represent the object. GPIS uses Gaussian process regression to approximate a closed SDF from observed data. Given a query point $\bm{x}$ in world coordinates, we can compute the expected mean and variance $(d_\mu(\bm{x}), d_\sigma(\bm{x}))$ of the signed distance $d$ between $\bm{x}$ and the object surface. The probability density of $\bm{x}$ to be on the object surface can be expressed as:
\begin{equation}
p(d(\bm{x})=0|\bm{x}) = \frac{1}{\sqrt{2\pi d_\sigma(\bm{x})^2}}e^{-\frac{d_\mu(\bm{x})^2}{2d_\sigma(\bm{x})^2}}
\end{equation}
When such surface uncertainty is present, it is hard to enforce initial contact to happen exactly at $\bm{p}_i(t_0)$ without making unintended contacts that cause the movement of the object. To mitigate this issue, our method extends the grasping process backward in time to model a pregrasp that places fingertips outside of the object surface before contact. The pregrasp can be represented as fingertip locations in the Cartesian space or a hand pose in the joint configuration space. We choose the latter because defining decision variables in the joint configuration space allows us to enforce hand kinematic constraints without the complexity of differentiating through an inverse kinematic process. Therefore, \textcolor{black}{following definition in \cite{preshape, pregrasp}}, we define a pregrasp $\bm{q}_\text{init} \in \mathcal{R}^{22}$ as a wrist 6D pose and joint angles of all fingers \textcolor{black}{such that fingers are not in contact with the object.} We assume that a fingertip location $f_\text{FK}(\bm{q}_\text{init}, i)$ can be linearly extrapolated from the vector $\bm{o}_i - \bm{p}_i(t_0)$ with the extrapolation coefficient $c\in(0,1]$:
\begin{equation}
\label{eqn:pregrasp_extrapolate}
 \bm{p}_i(t_0) = c(f_\text{FK}(\bm{q}_\text{init}, i) - \bm{o}_i) + \bm{o}_i,
\end{equation}
where the extrapolation coefficient $c$ adjusts the relative distance between surface to the target $\bm{o}_i$ and the distance between the surface to the pregrasp fingertip. When $c$ is larger, the pregrasp fingertips are closer to the surface. We determine $c$ manually $c = 0.7$ for all our experiments.

\begin{figure}[t]
  \centering
  \includegraphics[width=\linewidth]{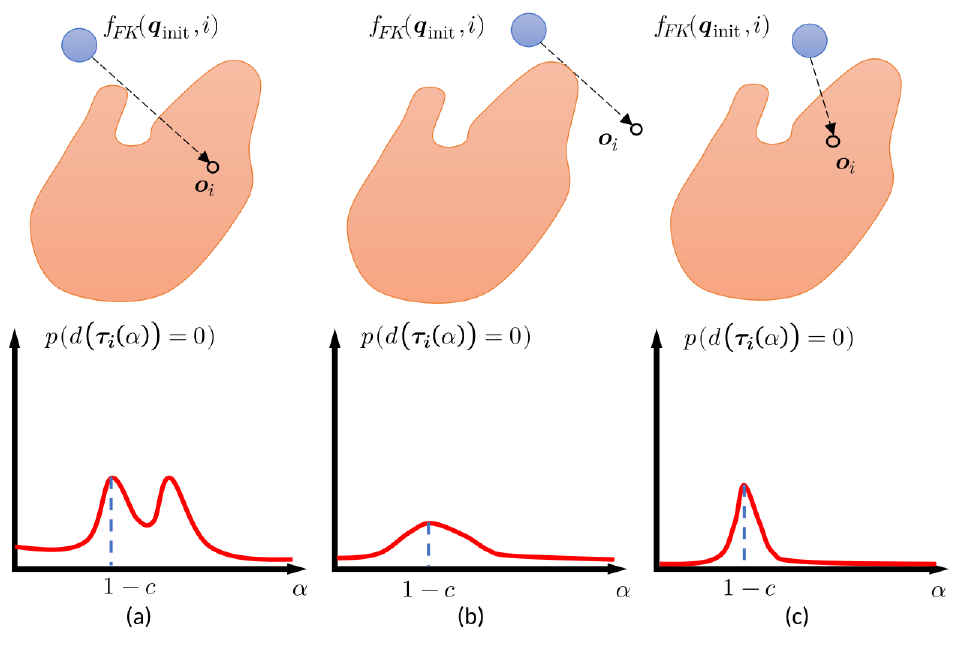}
  \caption{Shape of p.d.f. for different line trajectory, we prefer (c) instead of (a) and (b) as contact is mostly likely to happen at the expected time and location.}
  \label{fig:pdf}
\end{figure}

A desired pregrasp should result in contact locations that have high probability to be the on the surface of the object. In addition, the contact location should have a prominent probability comparing to points in its neighborhood as we shown in Fig.~\ref{fig:pdf}. We define a line segment $\tau_i(\alpha), \alpha \in [0,1]$ for each contact neighborhood, where $\tau_i(0) = f_\text{FK}(\bm{q}_\text{init}, i)$, $\tau_i(1) = \bm{o}_i$, and $\tau(1-c) = \bm{p}_i(t_0)$. We then define an objective term that encourages a desired pregrasp:
\begin{equation}
E_\text{uncer} = \sum_i^m\int_{0}^{1}  p(d(\bm{\tau}_i(\alpha))=0) - p(d(\bm{\tau}_i(1-c))=0)d\alpha
\end{equation}
To avoid computing integration, we use $K$ equally spaced sample points along the trajectory to approximate p.d.f and the energy term is computed as:
\begin{equation}
\label{eqn:uncer}
E_\text{uncer} = \sum_i^m\sum_{k}^K p(d(\bm{\tau}_i(\Delta \alpha k))=0) - p(d(\bm{\tau}_i(1-c))=0)
\end{equation}

\subsection{Optimizing compliant grasp with uncertain object surface}
With the introduction of pregrasp, $\bm{q}_\textbf{init}$, we augment the decision variables of the optimization to be $\bm{q}_\textbf{init}$, $\bm{p}_i(t_0)$, $\bm{o}_i$, and $k_i$, where $i\in \{1,2,...,m\}$. In practice, we drop $\bm{p}_i(t_0)$ since it can be expressed as a function of $\bm{q}_\text{init}$ and $\bm{o}_i$ (Equation \ref{eqn:pregrasp_extrapolate}). 

In addition to $E_{sp}$, $E_{dist}$, and $E_{uncer}$, we introduce four more objective terms to improve the quality of the compliant grasp.

\subsubsection{Low gains}
To further increase the compliance of the grasp, we encourage the controller gains to be as low as possible:
\begin{equation}
    E_\text{gain} = \sum_i^m k_i^2
\end{equation}


\subsubsection{Target position}
Ideally, the target location $\bm{o}_i$ should locate inside the object to encourage stability at equilibrium.
\begin{equation}
E_\text{tar} = \sum_i^m d(\bm{o}_i)
\end{equation}

\subsubsection{Collision avoidance}
To ensure the grasp is collision free, we penalize collision between fingers, between the hand and the object, and between the hand and the table. We approximate collision geometry of the hand using $16$ spheres with the centers defined as $\{\bm{s}_1, \cdots, \bm{s}_{16}\}$ in the world frame which can be evaluated by $f_\text{FK}(\bm{q}_\text{init})$ analytically. The radii of the spheres are defined manually as $\{r_1, \cdots, r_{16}\}$ (Details in Appendix \ref{sec:collisions}). For each self-collision pair $(i,j)$ the collision penalty is defined as: 
\begin{equation}
E_\text{self}^\text{i,j} = \begin{cases}
    \frac{1}{\text{dist}(\bm{s}_i, \bm{s}_j)} & \text{if dist}(\bm{s}_i, \bm{s}_j) \le r_i+r_j\\
    0 $ \text{otherwise}$
    \end{cases}
\end{equation}

The collision between the hand and the object is penalized by
\begin{equation}
E_\text{ho} = \sum_i\begin{cases}
                        \frac{1}{d(\bm{s}_i)}& \text{if }d(\bm{s}_i)\le r_i\\
                        0& \text{otherwise,}
                    \end{cases}
\end{equation}
and between the hand and the table by
\begin{equation}
E_\text{ht} = \sum_i\begin{cases}\frac{1}{\bm{s}_i^z}& \text{if }\bm{s}_i^z\le r_i\\
                                 0& \text{otherwise,}\end{cases}
\end{equation}
where $\bm{s}_i^z$ indicates the height of the sphere. Putting it together, the collision penalty term is defined as 
\begin{equation}
\label{eqn:col}
E_\text{col} = \sum_{i,j}^\text{collision pairs}E_\text{self}^\text{i,j} + E_\text{ho} + E_\text{ht}, 
\end{equation}
where collision pairs are defined in Appendix \ref{sec:collisions}. \textcolor{black}{As most collision spheres have the same radius, we don't normalize the energy for radius.}

\subsubsection{Regularization}
We also add some regularization to shape the optimization. Less object movement during the compliant grasping process is encouraged by minimizing the difference between $\bm{p}_i(t_0)$ and $\bm{p}_i(t_{eq})$. We also regularize the joint angles of $\bm{q}_\text{init}$ to match a neutral reference pose $\bm{q}_\text{ref}$ designed manually. To prevent degenerate solution with zero force at each contact point at equilibrium, we also encourage a minimal force exerted by each fingertip: 
\begin{equation}
\label{eqn:reg}
\begin{split}
E_\text{reg} = ||\bm{q}_\text{ref} - \bm{q}_\text{init}||_2^2 + \sum_i^m||\bm{p}_i(t_0) - \bm{p}_i(t_\text{eq})||_2^2\\
E_\text{force} = w_\text{force}\sum_i^m\min(f_\text{min},||\bm{f}_i||_2)
\end{split}
\end{equation}
where $f_\text{min} = 2$ and  $w_\text{force} = 200$.

Putting it all together, we arrive at our final optimization of a compliant grasp that aims to grasp an object with uncertain shape:
\begin{equation}
\label{eqn:energy}
\begin{split}
\underset{\bm{q}_\text{init}, \{\bm{o}\}, \{k\}}{\arg\min}\
     w_\text{sp}E_\text{sp} + w_\text{dist}E_\text{dist} + w_\text{uncer}E_\text{uncer} + w_\text{gain}E_\text{gain}\\
    + w_\text{tar}E_\text{tar} +w_\text{col}E_\text{col}+w_\text{reg}E_\text{reg}+w_\text{force}E_\text{force}
\end{split}
\end{equation}

\begin{table}[t]
    \centering
    \resizebox{\linewidth}{!}{
    \begin{tabular}{@{}cccccccc@{}}
    \toprule
    $w_\text{sp}$ & $w_\text{dist}$& $w_\text{uncer}$& $w_\text{gain}$ & $w_\text{tar}$ & $w_\text{col}$ & $w_\text{reg}$ & $w_\text{force}$\\
    \hline
    200 & 10000& 20 & 0.5 & 1000 & 1 & 10 & 200 \\
     \bottomrule
    \end{tabular}
    }
    \caption{Weights of different energy terms in Eq~\ref{eqn:energy}}
    \label{tab:weights}
\end{table}
The weight of each object term is shown in Table \ref{tab:weights}. The final objective function is fully differentiable and is optimized with off-the-shelf gradient-based optimizers.

\section{Experiments}
\begin{figure*}[t]
  \centering
  \includegraphics[width=\linewidth]{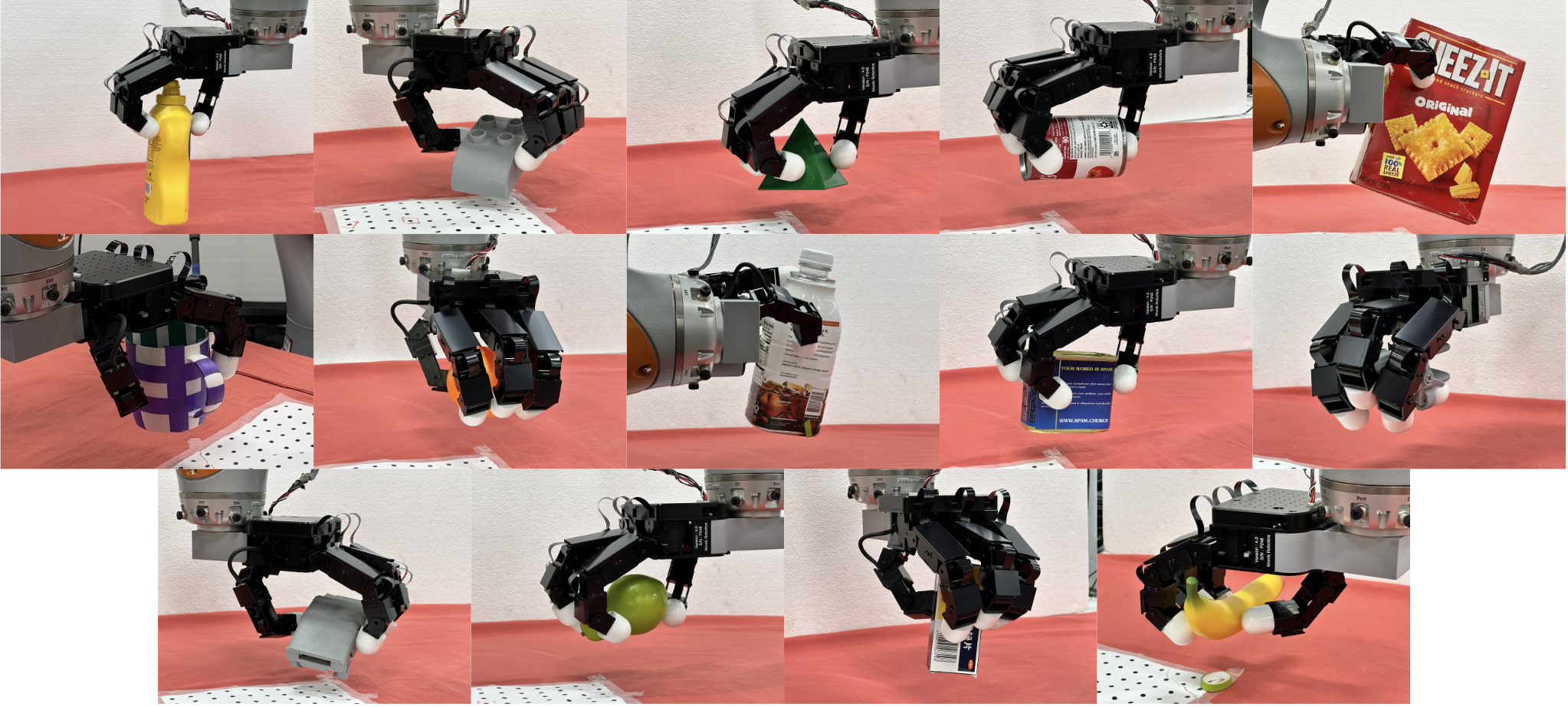}
  \caption{Grasping different object with single depth image input}
  \label{fig:result_vis}
\end{figure*}

In our experiment, we will demonstrate that optimizing controller gains is important for grasping objects under shape uncertainty. To this end, we compare the proposed approach to a baseline that uses bilevel optimization with the force closure criterion~(similar to \cite{dexgraspnet, diffwc}). For this baseline, the controller gains have to be picked and it does not consider object shape uncertainty. 
We also systematically study the impact of different levels of shape uncertainty on the performance of our approach. Additionally, we perform an ablation of the objective function with and without considering object shape uncertainty. Our experiments demonstrate that our approach outperforms the baselines by 18-27\% and that not taking uncertainty into account in the objective function leads to a significant performance drop (9\%).
Through an additional ablation, we will show that the impact of the pregrasp formulation on grasp performance is significant. Finally, we discuss the property of allowing object movement in the SpringGrasp metric and how it is beneficial for grasp planning.
Fig. \ref{fig:result_vis} shows an example grasp per object in our dataset.

\subsection{Evaluation metric}
Unless otherwise states, we use grasp success rate in real robot experiment as the evaluation metric for all our experiments. To evaluate whether a grasp is successful or not, the object is lifted 5cm after grasping. We use two criteria to assess the success of a grasp: 1) Whether the object is lifted. 2) Whether there is sliding between fingertips and the object. If a grasp satisfies both criteria, it is considered a successful grasp. If the object is lifted but slides during the process, the grasp is considered as partially successful. For each object, we attempt five grasps over different object poses. We then compute the grasp success rate as $\frac{\#\text{successful} + 0.5\#\text{partial}}{5}$. We also report the average success rate over all objects.

\subsection{Experiment setup}
\subsubsection{Hardware setup}
\begin{figure}[t]
  \centering
  \includegraphics[width=\linewidth]{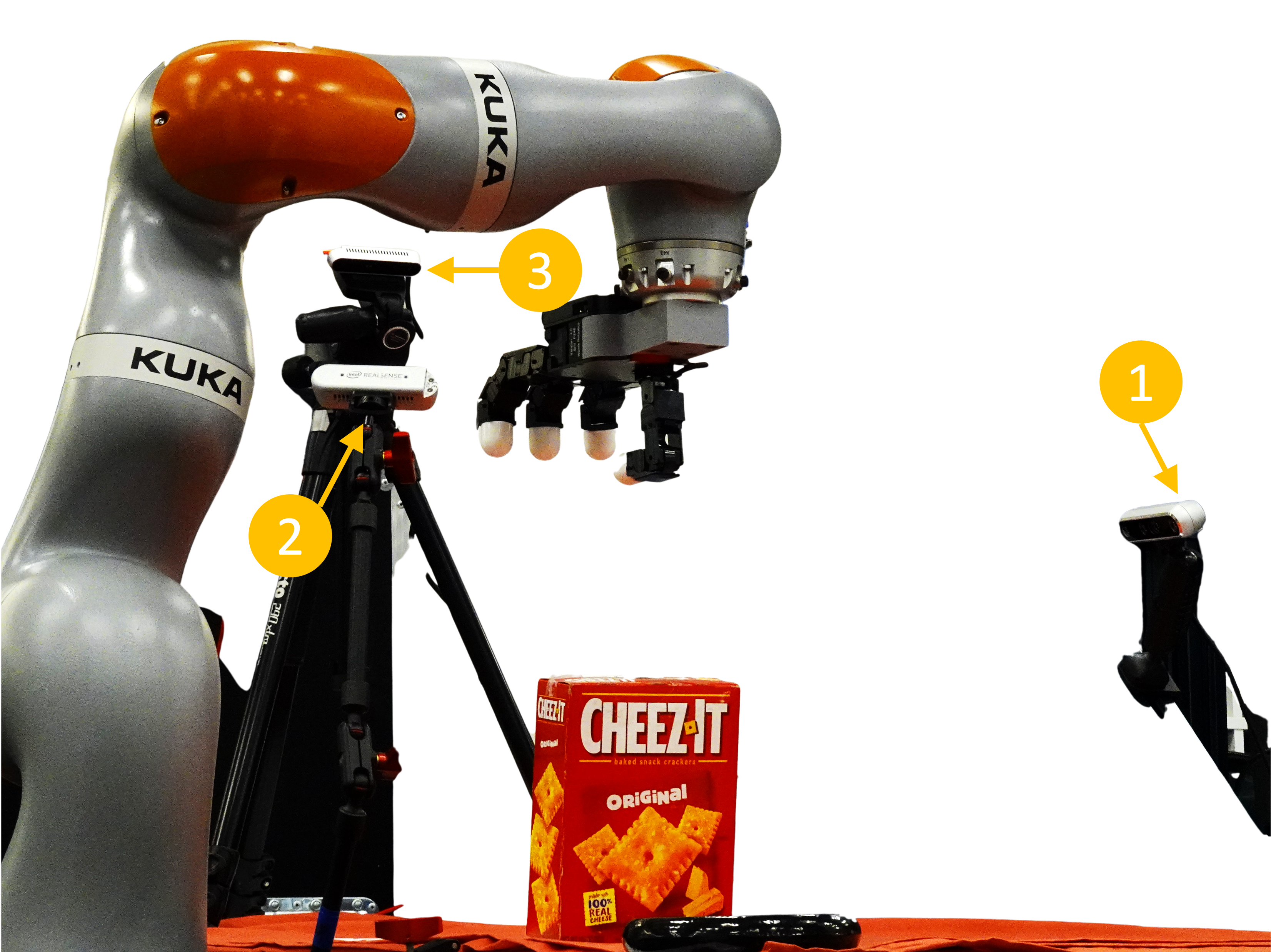}
  \caption{Real robot setup showing the Kuka iiwa arm equipped with a left Allegro hand and the configuration of the three RGB-D cameras.}
  \label{fig:setup}
\end{figure}

We use three realsense RGB-D cameras located around the table (see Fig. \ref{fig:setup}). 
The robot arm is Kuka iiwa R820 and is equipped with a left Allegro hand. All the devices are connected to a PC with a i7-13700K CPU and RTX3090 GPU, which is also used for grasp optimization and motion planning for the arm.

\begin{figure}[t]
  \centering
  \includegraphics[width=\linewidth]{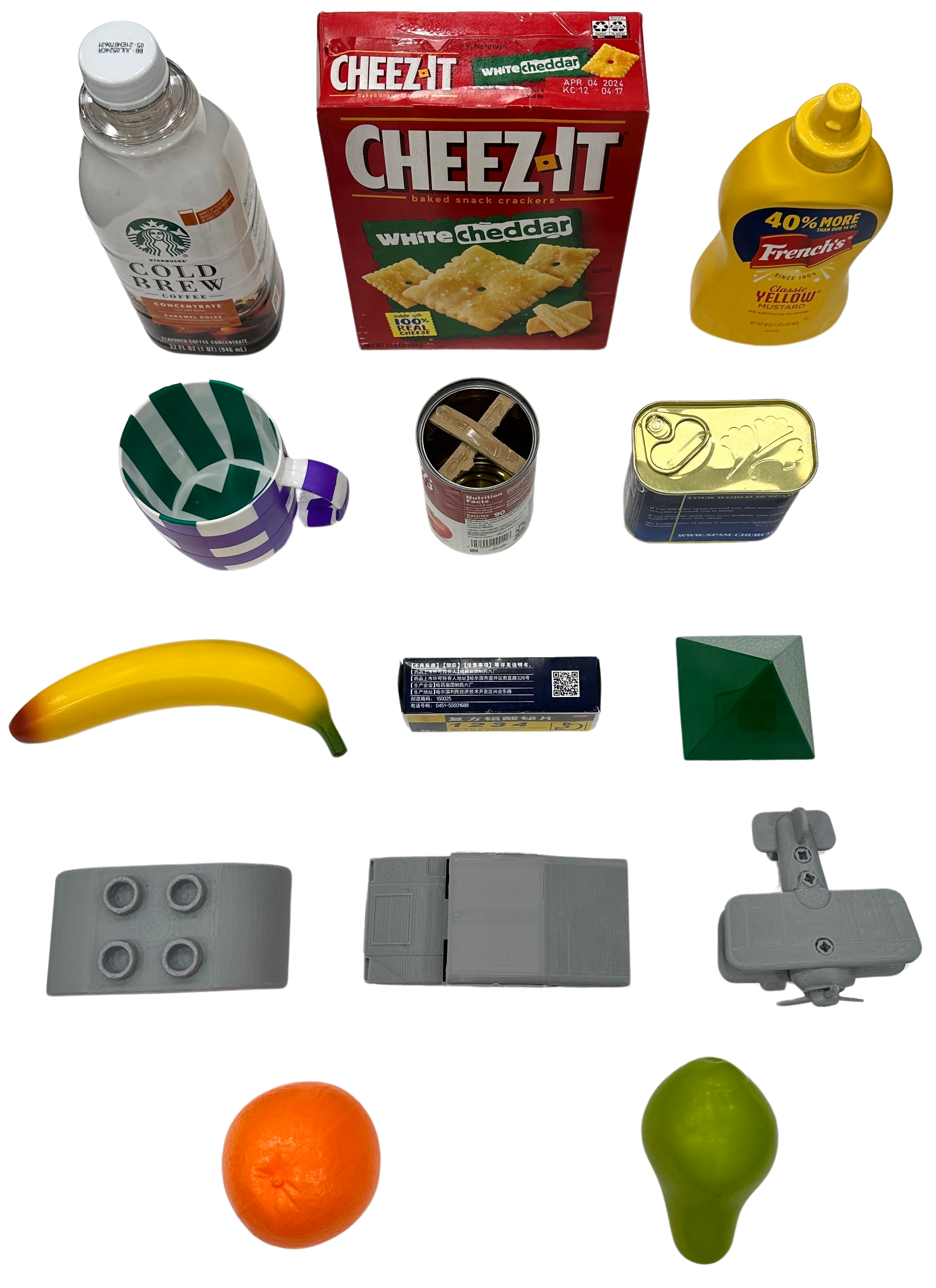}
  \caption{All objects used for experiments}
  \label{fig:objects}
\end{figure}

\subsubsection{Software setup}
For robot control, our Allegro hand impedance controller is adapted from the AllegroHand KDL \cite{allegrokdl} and our Kuka Arm controller is adapted from iiwaPy \cite{iiwapy}. Our SpringGrasp planner is written in Pytorch and we solve the optimization using RMSProp. After obtaining a grasp, we use curobo \cite{curobo} to plan a collision free arm trajectory to the desired wrist pose. 

\subsection{Baseline}
For our baseline, we use bilevel optimization with a differentiable force closure energy term similar to \cite{dexgraspnet, diffwc}. We use this baseline to show the advantage of optimizing the controller gains with the SpringGrasp metric compared to force closure bilevel optimization where the user has to manually tune the gains.
The energy function of this baseline is adapted from our method with $E_\text{sp}$ changed into the force closure metric energy $E_\text{fc}$. More details can be found in the Appendix. \ref{sec:baseline}. 
To directly compare the performance between using the SpringGrasp or the force closure metric, we compute the target location $\bm{o}_i$ with $i\in\{1,2,...,m\}$ where $m$ is the number of fingers and pregrasp hand pose $\bm{q}_\text{init}$ based on the force $\bm{F}_i(t_0)$ that is the result of the inner loop of the bilevel optimization problem and pick a controller gain $k_i$:
\[
\bm{o}_i = \bm{p}_i(t_0) + \frac{\bm{F}_i(t_0)}{k_i},i\in\{1,2,...,m\}
\]
\[
\bm{q}_\text{init} = \text{IK}(\frac{1}{c}\{\bm{p}_i\}- \frac{1-c}{c}\{\bm{o}_i\}), c\in[0,1]
\] 
$c_i$ are the same clearance coefficients as used in our method.

\subsection{Comparison with Baseline without Optimized Gains}
To demonstrate the importance of optimizing the controller gains in robotic grasping, we conducted experiments using two sets of gains in the baseline. We refer to these sets as either Bilevel (high), Bilevel (low) or Bilevel(heuristic). For Bilevel (high), we set the gains to $k_1, k_2, k_3 = 160, k_4 = 320$. For Bilevel (low), we used $k_1, k_2, k_3 = 80, k_4 = 160$, which is close to the average value output by our approach. \textcolor{black}{For Bilevel(heuristic), we set the gains inverse proportional to variance of expected contact location as $k_i = \frac{\alpha}{d_\sigma(\bm{p}_i)}$, which is similar to the setting in \cite{mostsimilar}. For this experiment, we estimate the object shape using point clouds collected from two camera views that are then input to GPIS.}

The result of the comparison between our method and the baseline is shown in Table \ref{tab:comparison}. With input point clouds from two viewpoints, our method achieved an overall grasp success rate of \textcolor{black}{89\%}, which is \textcolor{black}{24\%} higher than that of Bilevel (low) and \textcolor{black}{27\%} higher than Bilevel (high). \textcolor{black}{Compared to Bilevel (heuristic), our method also achieves an 18\% higher success rate.} The performance difference between our method and Bilevel (low) was more pronounced with long and thin objects, such as boxes and bananas. For instance, the success rate for grasping a banana dropped from 80\% to 40\%, and for a box from 100\% to 60\%. When compared with Bilevel (high), the difference was more significant with non-convex objects like bananas and cars. \textcolor{black}{Although both types of failures were mitigated in Bilevel (heuristic), there is still a significant success rate difference when grasping thin and long objects as well as non-convex objects.}

\begin{figure}[t]
  \centering
  \includegraphics[width=\linewidth]{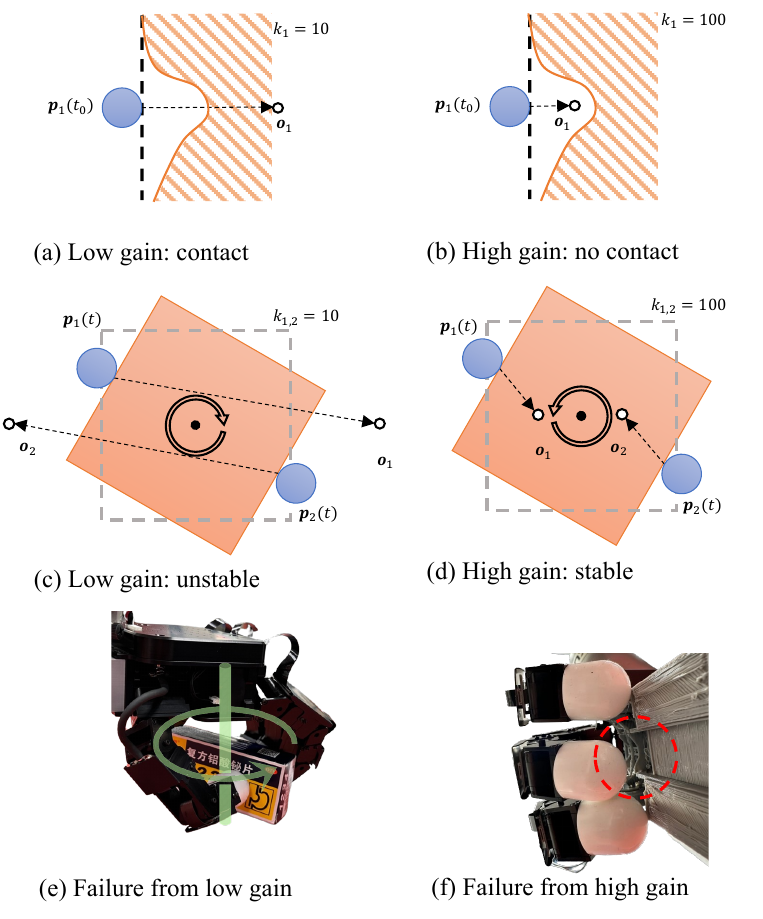}
  \caption{Illustration of baseline failure modes. The fingertip needs to apply the same force despite different controller gains (either $k_i=10$ for (a,c) or $k_i=100$ for (b,d)). For shaping uncertain object (top row - the dashed line indicates the expected object surface; the solid curve is the actual surface), using low gains $k_i$ can ensure making contact with the object as shown in (a), instead of not making contact as shown in (b). For smaller objects (middle row), when the object is perturbed (represented by the solid box) from its equilibrium (represented by the dashed box) at time $t$, using higher gains can ensure that the fingertip can always generate an object torque to compensate a perturbation as shown in (c) instead of continuing to twist the object in the direction of the perturbation as shown in (d). In real robot experiments (bottom row), the box was twisted in (e) due to an unstable equilibrium while in (f) the middle finger fails to attach to the object because the target location is outside the object.}
  \label{fig:stiffcompliant}
\end{figure}

The results suggest that there is no universal set of gains that works for all scenarios. This is illustrated in Fig.~\ref{fig:stiffcompliant}. On the one hand, low gains can result in an unstable equilibrium where even small perturbations are sufficient to spin the object out of the grasp. 
High gains can position the target locations $\bf{o}_i$ close to the zero level set of the GPIS. This can result in the finger not making contact with the true object surface that may be behind the estimated surface. SpringGrasp allows to also optimize the controller gains and therefore to avoid unstable equilibria or missed contacts.

\begin{table*}
\centering
\resizebox{\linewidth}{!}{
\begin{tabular}{l|ccc|c|c|cc>{\color{black}}c}
\toprule
\multicolumn{1}{c|}{} & \thead{Ours \\ (3 views)}        & \thead{Ours \\ (2 views)}       &  \thead{Ours \\ (1 view)}       & \thead{w/o uncertainty \\ (1 view)} & \thead{w/o pregrasp \\ (1 view)} & \thead{Bilevel-high \\ (2 views)} & \thead{Bilevel-low \\ (2 views)} & \thead{Bilevel-heu \\ (2 views)} \\ \hline
Mustard bottle        & 90\%           & \textbf{100\%} & 80\%          & 70\%             & 90\%          & 50\%           & 60\%    &    60\%       \\
Lego                  & 90\%           & 70\%          & 80\%          & 70\%             & 60\%          & 70\%           & 50\%     &    50\%      \\
Pyramid               & \textbf{100\%} & \textbf{100\%} & \textbf{100\%} & 80\%             & 70\%          & 70\%           & 70\%   &    80\%        \\
Campell can           & \textbf{100\%} & \textbf{100\%} & \textbf{100\%} & 90\%             & \textbf{100\%} & 90\%           & 90\%  &    80\%         \\
Cheezit box          & 80\%           & \textbf{90\%} & 80\%          & 60\%             & \textbf{90\%} & 30\%           & 40\%      &    80\%     \\
Mug                   & \color{black}{\textbf{90\%}}  & \color{black}{80\%}         & \color{black}{80}\%          & \color{black}{70\%}             & \color{black}{80\%}          & \color{black}{70\%}           & \color{black}{60\%}    &    \color{black}{80\%}       \\
Orange                & \textbf{100\%} & \textbf{100\%} & 90\%          & \textbf{100\%}   & \textbf{100\%} & 90\%           & 80\%   &    \textbf{100\%}        \\
Coffee bottle         & 80\%           & \textbf{90\%} & \textbf{90\%} & 70\%             & \textbf{90\%} & 50\%           & 60\%     &    60\%      \\
Spam                  & \textbf{80\%}  & 60\%          & 60\%          & 60\%             & 60\%          & 40\%           & 60\%     &    50\%      \\
Plane                 & \textbf{100\%} & 70\%          & 70\%          & 80\%             & 80\%          & 60\%           & 70\%     &    70\%      \\
Car                   & 90\%           & \textbf{100\%} & 70\%          & 70\%             & 70\%          & 60\%           & 80\%    &    60\%       \\
Banana                & \textbf{100\%} & 80\%          & 90\%          & 60\%             & 70\%          & 30\%           & 40\%     &    60\%      \\
Pear                  & \textbf{100\%} & \textbf{100\%} & 90\%          & \textbf{100\%}   & 80\%          & 80\%           & 80\%    &    80\%       \\
Small box             & \textbf{100\%} & \textbf{100\%} & \textbf{100\%} & 70\%             & 70\%          & 80\%           & 60\%   &    90\%        \\ \hline
\textbf{Total}        & \textbf{93\%}  & 89\%          & 84\%          & 75\%             & 79\%          & 62\%           & 65\%     &    71\%     \\ \bottomrule
\end{tabular}}
\caption{Grasp success rate over 5 trials per object. Note that we can have successful and partially successful grasps. {\em Ours\/} refers to SpringGrasp that takes as input point clouds from either 3, 2 or 1 viewpoint. {\em w/o uncertainty\/} and {\em w/o pregrasp\/} are ablations of our method when either removing $E_\text{uncer}$ or setting $c = 1$ respectively in Eq.~\ref{eqn:pregrasp_extrapolate}. Both use a single view point cloud as input. {\em Bilevel-high\/}, {\em Bilevel-low\/} and \textcolor{black}{\em Bilevel-heu\/} refer to the bilevel optimization baseline with high gains, low gains, \textcolor{black}{or heuristically selected gains.} They all use point clouds from two views as input. Our method significantly outperforms the baseline as well as the ablations even under maximum uncertainty in object shape.}
\label{tab:comparison}
\end{table*}

\subsection{Robustness to Varying Levels of Shape Uncertainty}
\begin{figure}[t]
  \centering
  \includegraphics[width=\linewidth]{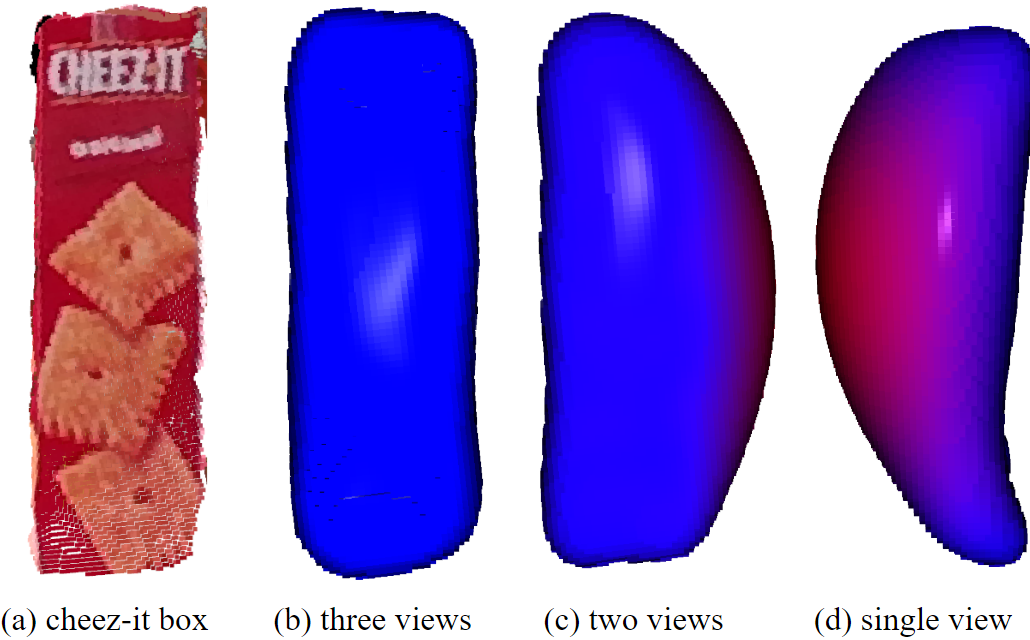}
  \caption{GPIS reconstructed surfaces from different number of viewpoints. The color on the surface indicates variance of a surface point where hotter colors correspond to a larger variance.}
  \label{fig:gpis_reconstruct}
\end{figure}

To study how robust our method is against varying levels of shape uncertainty, we evaluate grasp success rate of our method using point clouds recorded from either three, two or one viewpoint. For each setting, we pick the viewpoint that provides the most information about the object (Camera 1 and 2 for experiments with two viewpoints and Camera 3 for single viewpoint experiments). As shown in Tab. \ref{tab:comparison}, grasp success rate slightly drop from 91\% to 89\% when changing from three to two viewpoints. The success rate drops to 84\% when only using point cloud from a single viewpoint. Contradict to our expectation, for larger object such as the Cheezit box, the grasp success rate does not drop significantly when changing from two to one viewpoint. As shown in Fig. \ref{fig:gpis_reconstruct}, GPIS estimates the true shape of the object fairly well een from one viewpoint and preserves necessary geometry information for successful grasping. For smaller objects, our method is quite robust to a reduction of information.

\subsection{Ablation study}
\begin{figure}[t]
  \centering
  \includegraphics[width=\linewidth]{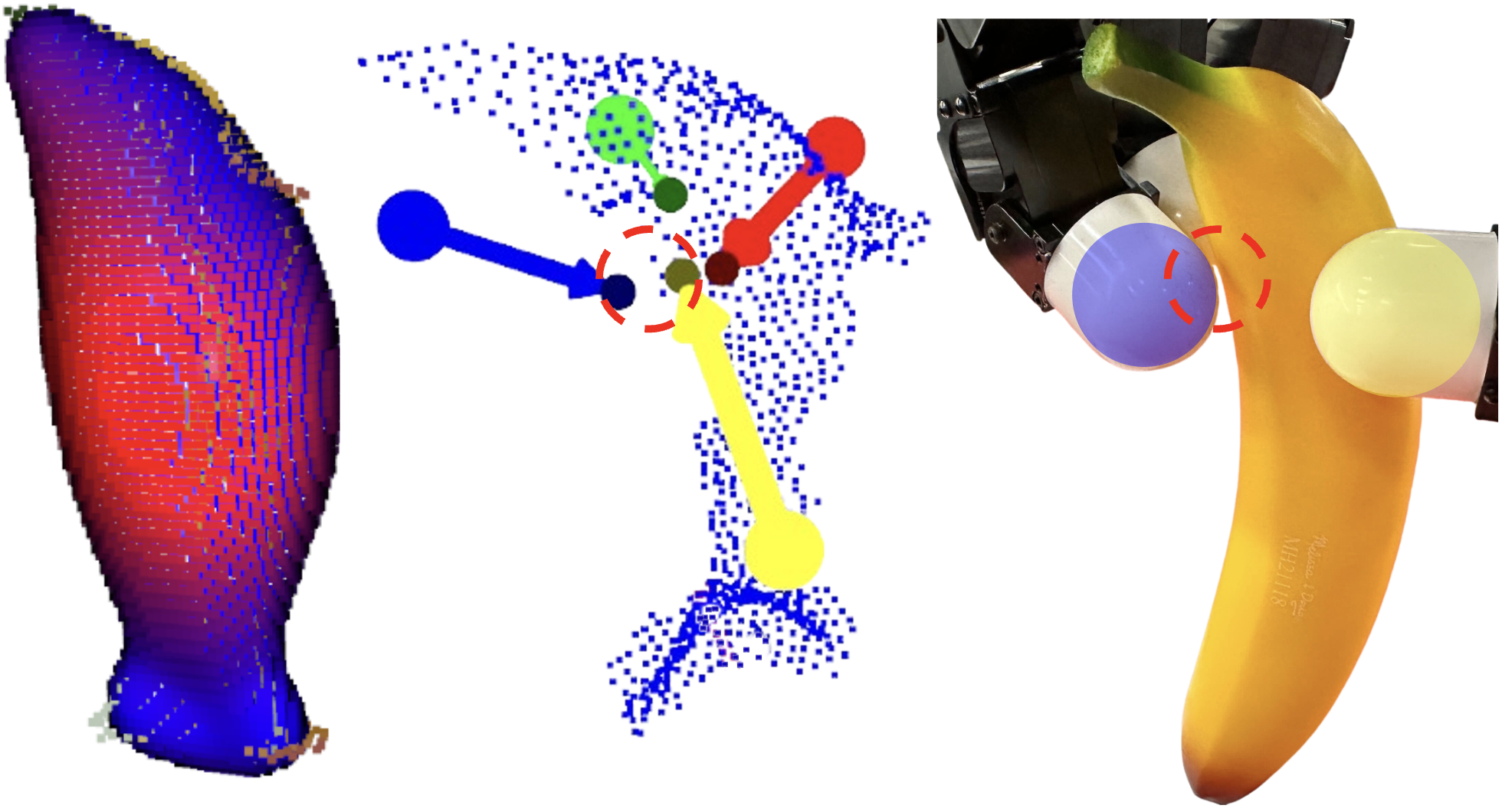}
  \caption{The left figure illustrates the uncertain surface modeled by GPIS. The central figure depicts the anticipated trajectory of the fingertips in the predicted pregrasp. The right figure shows that the tip of the ring finger (blue) does not make contact with the actual object surface, resulting in a partial grasp failure. For visualization, we color visible thumb and ring finger in the right figure to match the colors in the middle figure.}
  \label{fig:uncertainty}
\end{figure}

In our method, we hypothesize that both considering uncertainty and using a pregrasp is essential to achieve a successful grasp under object shape uncertainty. We justify our design with an ablation study on these two components. To understand the importance of the term $E_\text{uncer}$ in Eq.~\ref{eqn:uncer}, we evaluate the grasp success rate on all objects with or without this term in the objective function. Note that for this experiment, we assume as input an object point cloud from a single view which is the most common scenario for robotic grasping. As shown in Tab. \ref{tab:comparison}, the grasp success rate drops from 84\% to 75\% when removing $E_\text{uncer}$ from Eq.~\ref{eqn:uncer} demonstrating the effectiveness of considering uncertainty in our framework. We also observed a typical failure mode when not considering uncertainty visualized in Fig. \ref{fig:uncertainty}. The optimized grasp may make contact with regions on the GPIS surface that have high uncertainty and therefore tend to be far off from the true object surface. This scenario results in missing or unexpected contacts which adversely affects grasp success rate. 

\begin{figure}[t]
  \centering
  \includegraphics[width=\linewidth]{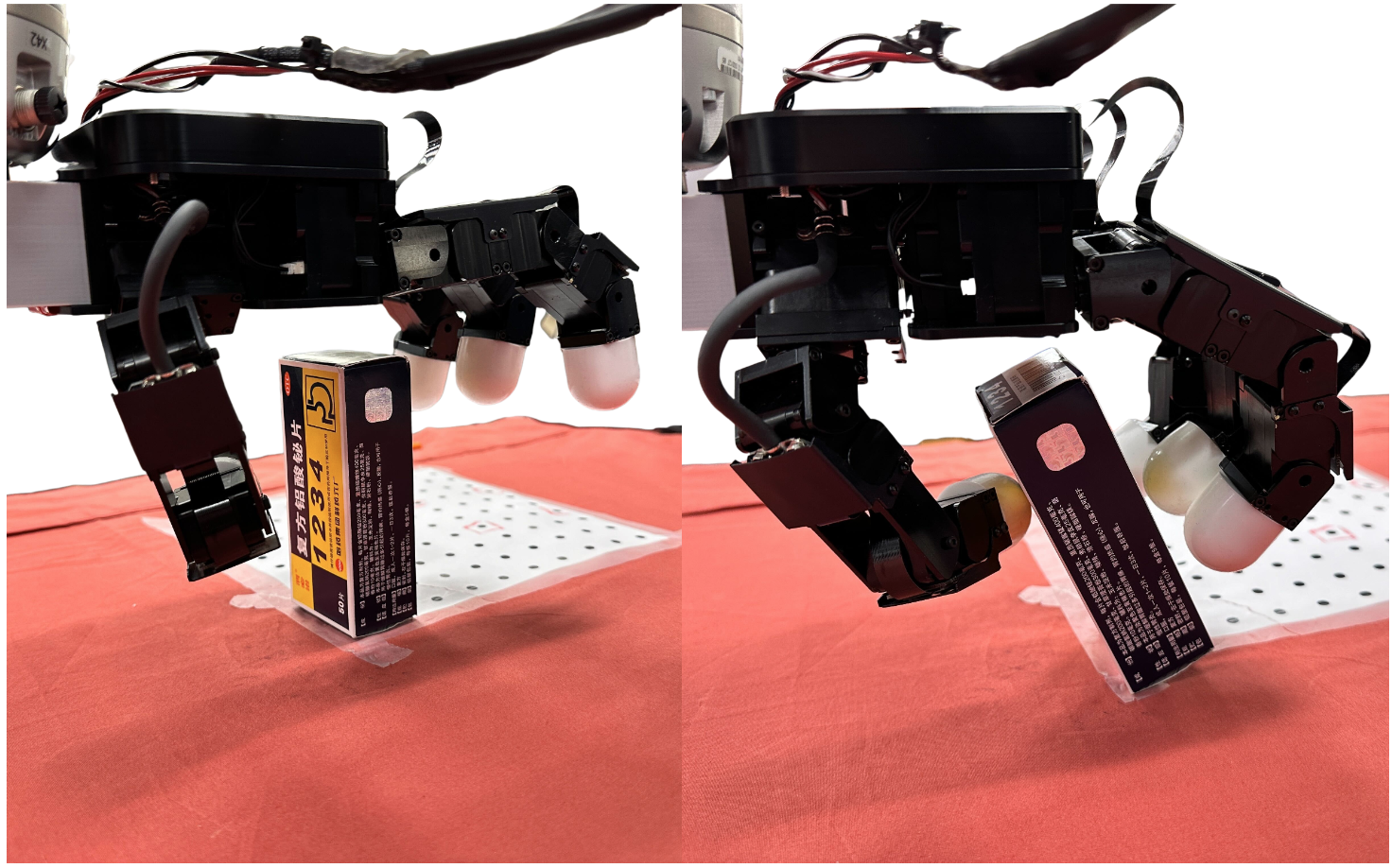}
  \caption{Grasp failure when not using pregrasp optimization. The object tips before the optimized grasp can be acquired.}
  \label{fig:pregrasp}
\end{figure}

To analyse the benefit of optimizing a pregrasp, we compare the grasp success rate of our method with different pregrasp coefficients $c = 1.0$ and $c = 0.7$. A pregrasp becomes a grasp (i.e. the fingertips are in contact with the estimated object surface) if the pregrasp coefficient $c=1.0$. From Tab.~\ref{tab:comparison}, we found that the grasp success rate decreases from 84\% to 78\% if we don't optimize for a pregrasp in which the fingertips are slightly offset from the estimated object surface. Among all objects and poses, the grasp success rate change most significantly if the object is not at a stable pose before grasping. In this case, it is most vulnerable to being perturbed by unexpected contact when the hand is reaching for the grasp pose. Fig. \ref{fig:pregrasp} shows an example where the box is standing on its side and is perturbed by premature contact resulting in it tipping over before the optimized grasp can be acquired.

\textcolor{black}{\subsection{Hyperparameter analysis}
Tab.~\ref{tab:weights} lists the weights we used for the energy terms in Equation \ref{eqn:energy}. Here we study how the selection of different weights affects the performance of our grasp planner. For each weight, we scale it by two multipliers, 0.5 and 2, and measure the grasp success rate respectively. We keep all other parameters unchanged when scaling the parameter of interest. We experiment with three distinctive objects, a pear, mustard bottle, and a Cheez-it box. The results shown in Tab.~\ref{tab:gridsearch} show that in general changing weights of different energy terms moderately will not cause significant changes in the grasp success rate, though some parameters are more sensitive than others such as $w_\text{tar}$ and $w_\text{reg}$.}

\begin{table}[t]
\centering
\resizebox{\linewidth}{!}{
\begin{tabular}{l|rrrrrrrr}
\hline
{\color{black} multiplier} & \multicolumn{1}{l}{{\color{black} $w_\text{sp}$}} & \multicolumn{1}{l}{{\color{black} $w_\text{dist}$}} & \multicolumn{1}{l}{{\color{black} $w_\text{uncer}$}} & \multicolumn{1}{l}{{\color{black} $w_\text{gain}$}} & \multicolumn{1}{l}{{\color{black} $w_\text{tar}$}} & \multicolumn{1}{l}{{\color{black} $w_\text{col}$}} & \multicolumn{1}{l}{{\color{black} $w_\text{reg}$}} & \multicolumn{1}{l}{{\color{black} $w_\text{force}$}} \\ \hline
{\color{black} 0.5} & {\color{black} 0.80} & {\color{black} \textbf{0.83}} & {\color{black} 0.77} & {\color{black} 0.77} & {\color{black} 0.73} & {\color{black} \textbf{0.93}} & {\color{black} 0.77} & {\color{black} \textbf{0.90}}\\
{\color{black} 1}   & {\color{black}0.83} & {\color{black} \textbf{0.83}} & {\color{black} \textbf{0.83}} & {\color{black} \textbf{0.83}} & {\color{black} \textbf{0.83}} & {\color{black} 0.83} & {\color{black} 0.83} & {\color{black} 0.83}\\
{\color{black} 2}   & {\color{black} \textbf{0.90}} & {\color{black} 0.77} & {\color{black} \textbf{0.83}} & {\color{black} \textbf{0.83}} & {\color{black} 0.77} & {\color{black} 0.83} & {\color{black} \textbf{0.90}} & {\color{black} 0.83}\\ \hline
\end{tabular}}
\caption{\textcolor{black}{Grasp success rate with different scaling factors on different weights}}
\label{tab:gridsearch}
\end{table}
\textcolor{black}{\subsection{Computation time}
We report the time taken to optimize grasps from 7 initial guesses and single initial guesses for both our method and baseline on both CPU and GPU in Tab.~\ref{tab:time}. Each time is measured as an average of 5 experiments. In general, our method optimizes faster than the bilevel baseline. When scaling up the number of seeds, the main computation bottleneck is forward kinematics computation provided by \cite{diffrobot}. Notice that due to implementation limitations in differentiable robot model\cite{diffrobot}, grasp planning runs faster on CPU than GPU.}
\begin{table}[t]
\centering
\resizebox{0.8\linewidth}{!}{
\begin{tabular}{l|rrrr}
\hline
{\color{black} time(s)} & \multicolumn{1}{l}{{\color{black} Ours (7)}} & \multicolumn{1}{l}{{\color{black} Ours (1)}} & \multicolumn{1}{l}{{\color{black} Bilevel (7)}} & \multicolumn{1}{l}{{\color{black} Bilevel (1)}} \\ \hline
{\color{black} CPU}     & {\color{black} 15.58}                       & {\color{black} 7.02}                       & {\color{black} 34.97}                          & {\color{black} 9.17}                          \\
{\color{black} GPU}     & {\color{black} 23.44}              & {\color{black} 14.22}              & {\color{black} 41.47}                          & {\color{black} 15.28}                          \\ \hline
\end{tabular}}
\caption{\textcolor{black}{Optimization time in second. Results of our methods and bilevel baseline using 1 and 7 seeds are represented as Ours (1), Ours (7), Bilevel (1), and Bilevel (7) respectively.}}
\label{tab:time}
\end{table}
\subsection{The Benefit of Modeling Object Movement during Grasp Acquisition} 
\label{subsec:coverage}
\begin{figure}[t]
  \centering
  \includegraphics[width=\linewidth]{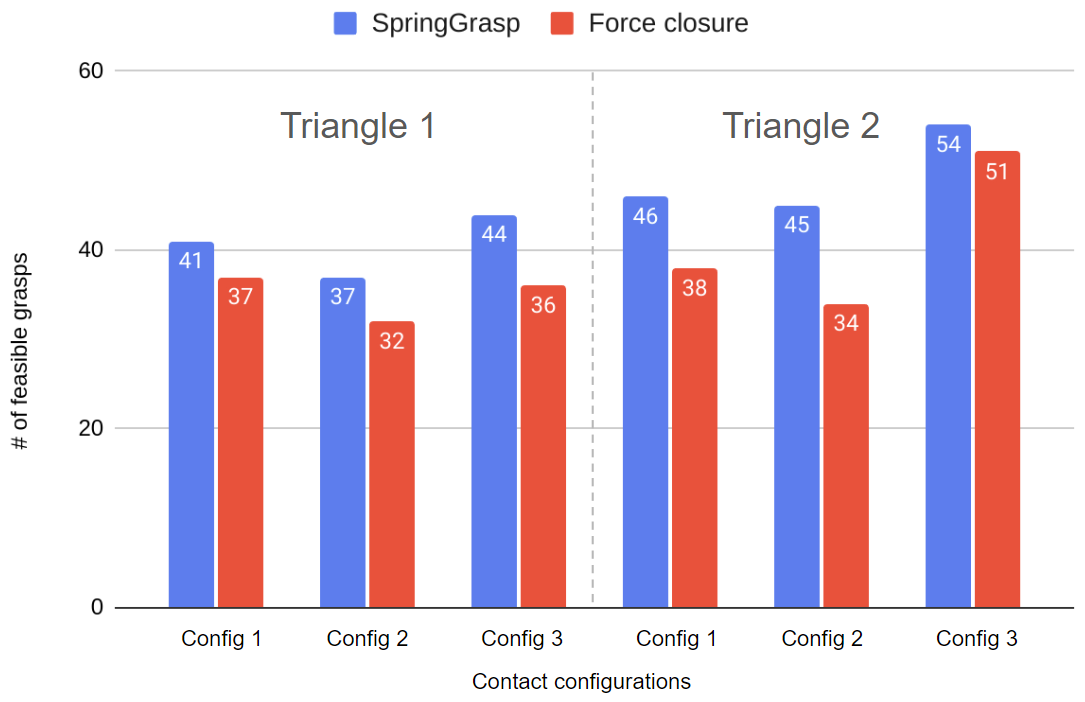}
  \caption{ (Blue) Number of feasible grasps generated by our method for 100 object poses per contact configuration. (Red) Number of feasible force closure grasps that do not require object movement for 100 object poses per contact configuration. 
  }
  \label{fig:coverage}
\end{figure}
\begin{figure}[t]
  \centering
  \includegraphics[width=\linewidth]{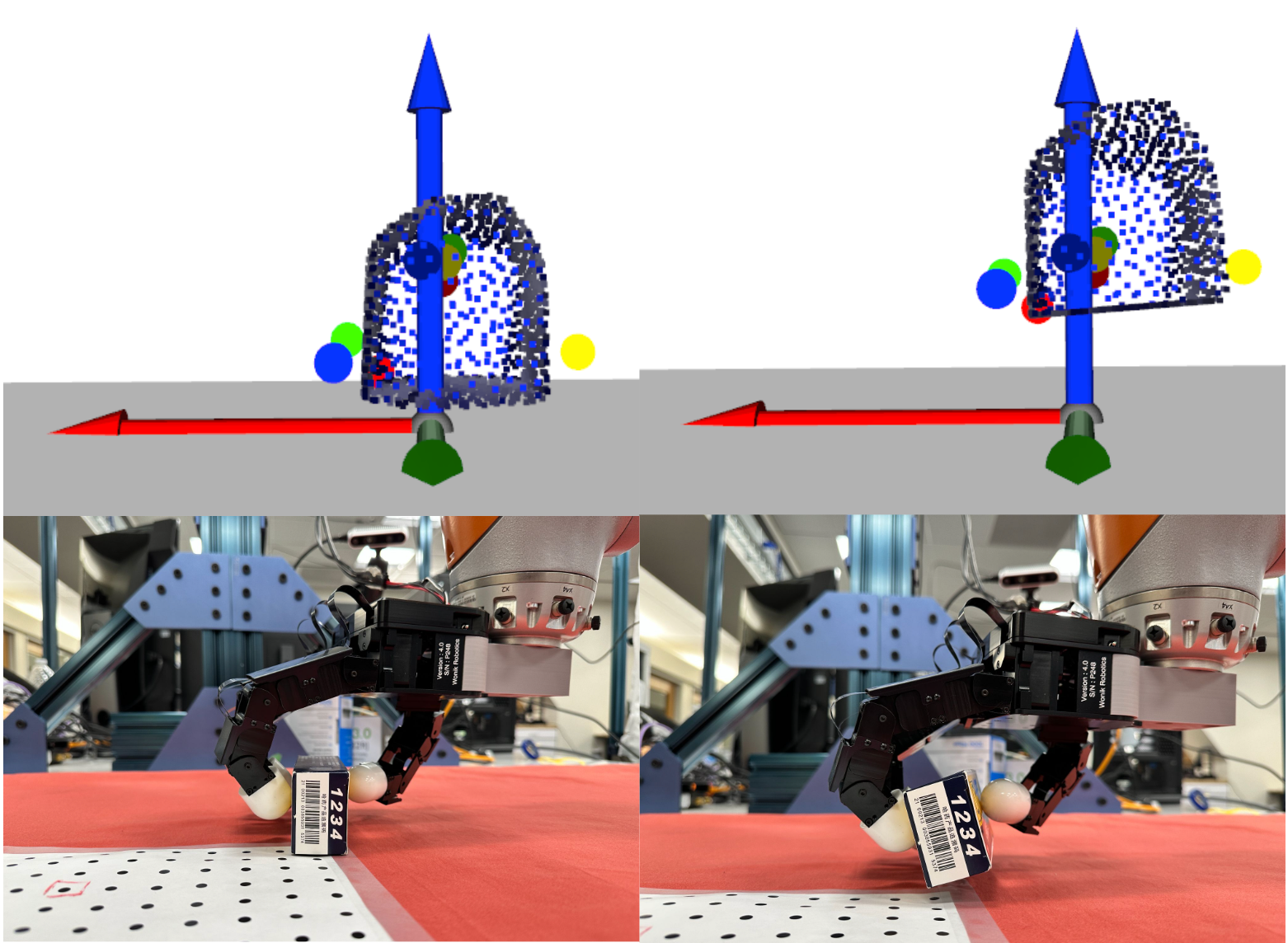}
  \caption{Object move up by 1cm as anticipated during grasp planning.}
  \label{fig:move}
\end{figure}

An additional property of the proposed SpringGrasp metric is that we model grasping as a dynamic process and allow object movement during grasping \textcolor{black}{as we shown in Sec.\ref{subsec:formulation}}. In this section, we will present evidence that allowing object movement can help to get around kinematic and joint torque limits when directly trying to establish a force closure grasp. For example, if a finger cannot apply the force required by the optimizer due to the current joint configuration and joint torque limits, the robot could first move the object into a new pose that leads to a joint configuration which allows a stable grasp. To test this hypothesis, we setup a 2D simulated experiment with two 2D triangles that are grasped by a three-fingered hand. For each triangle, we generate three different contact configurations on the triangle surface and sample 100 triangle poses per triangle and contact configuration. In each experiment, we optimize grasps with our metric as follows: as in this experiment, we only care about kinematic constraints and joint torque limits, we optimize target locations $\bm{o}_i$ and gains $k_i$ while contact locations $\bm{p}_i$ remain the same according to the three contact configurations we picked. Note that the wrist of the hand remains at a fixed location. Of the generated grasps, we only keep those that are feasible and can apply the desired force (with or without triangle movement) as verified in simulation. Of those feasible grasps, we evaluate how many of them are in force closure and in which each finger can apply enough force under joint torque constraints without moving the triangle (more details in Appendix. \ref{sec:sim_exp}). Grasps that allow the fingers to directly achieve force closure without triangle movement are only a subset of all feasible grasps. This means that there is no feasible grasp that could form force closure without triangle movement but fails to obtain a feasible compliant grasp. From Fig. \ref{fig:coverage}, only around 85\% of feasible grasps can directly achieve force closure without triangle movement. This demonstrates that there is a significant portion (15\%) of object poses that can benefit from object movement to avoid such kinematic and joint torque limits. Using the SpringGrasp metric, we can also deliberately control the object movement during optimization. Using an extra energy term $E_\text{pose} = \sum_i ||\bm{p}_i^\text{des}(t_\text{eq}) - \bm{p}_i(t_\text{eq})||_2$ to replace $\sum_i ||\bm{p}_i(t_\text{0}) - \bm{p}_i(t_\text{eq})||_2$ in Eq.~\ref{eqn:reg}, we can encourage object and fingertips to move in some desired direction, such as up by 1cm. For example, in Fig. \ref{fig:move} we set $\bm{p}_i^\text{des}(t_\text{eq}) = \bm{p}_i(t_0) + [0, 0, 0.01]$ which could be used to minimize undesired contact with the table during grasping.

\section{Conclusion \& future work}
We present an optimization-based system that can generate pregrasp grasps from noisy and partial point cloud observation. Benefiting from the SpringGrasp metric, pregrasp formulation, and consideration of uncertainty, our method could grasp a diverse set of objects with an 89\% success rate from 2 views and 84\% grasp success rate from single view in a real robot experiment. \textcolor{black}{When grasping objects with our method, the most common failure happens when the object being grasped is heavier than 0.3kg. This is shown in the experiment with the fully loaded spam can (0.4kg). As our method doesn't model gravity explicitly, its ability to grasp heavy objects is limited. A natural next step would be to integrate gravity and uncertainty into object mass into our grasp planner.} Moreover, the GPIS model suffers from modeling objects with thin walls such as knives or baskets due to the challenge of distinguishing points from the inner surface and outer surface and effectively sampling the interior point of the object. Future work could focus on decomposing the object geometry and extracting part of the object that is easy to grasp from the point cloud such as the handle of the knife and basket. \textcolor{black}{Another interesting direction could be using our grasp planner to generate a dataset and train a deep neural network to directly predict compliant grasp, given a partial observation of the object.}

\section{Acknowledgement}
This project is supported by NSF:FRR 2153854


\bibliographystyle{plainnat}
\bibliography{references}
\clearpage
\newpage
\appendix
\subsection{Analysis of contact margins}
\label{sec:apped_proof}
\begin{figure}[h]
  \centering
  \includegraphics[width=0.8\linewidth]{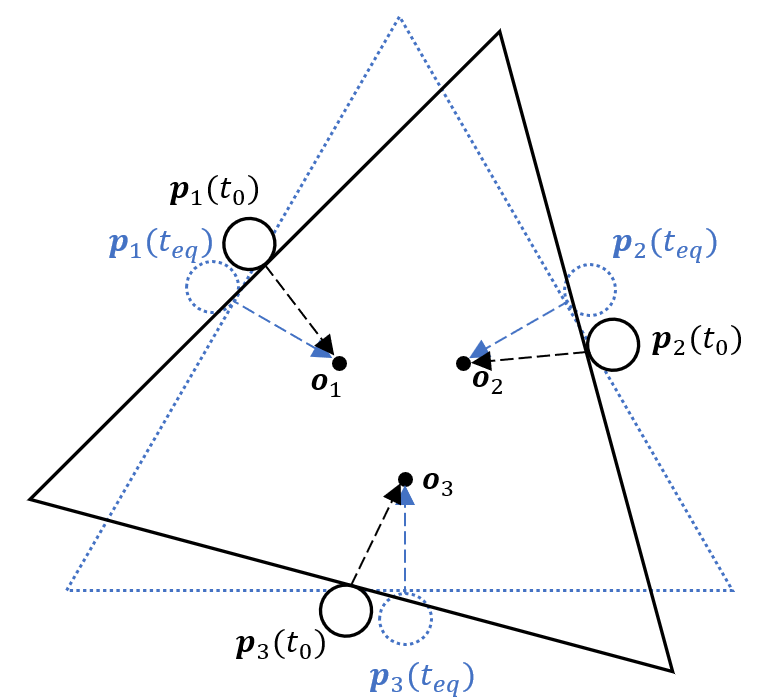}
  \caption{Three fingertips making contact with a 2D triangle. Fingertips apply force toward $\bm{o}_i$ and move with the object from $\bm{p}_i(t_0)$ to $\bm{p}_i(t_\text{eq})$}.
  \label{fig:example}
\end{figure}

This section intends to justify why the feasibility of the initial and target state according to the proposed SpringGrasp metric is a good heuristic to indicate feasibility of the entire dynamic process. For simplicity of the analysis, we consider a problem with three fingertips that make contact with a 2D object as shown in Figure \ref{fig:example}. In this system, the pose of the object can be described by $\bm{s}(t) = \{x(t),y(t),z(t),\theta(t)\}$. If the damping coefficients of each virtual spring are sufficiently large, no oscillation will happen during the dynamic process. This means that for any time, we can express it as convex combination of initial state and target state with a blending coefficient $\beta(t),t\in[t_0,t_\text{eq}]$ where $\beta(t_0)=0$ and $\beta(t_\text{eq})=1$:
\[
\bm{s}(t) = (1-\beta(t)) \bm{s}(t_0) + \beta(t)\bm{s}(t_\text{eq})
\]

\subsubsection{Translation only}
We define the angle between the force vector $\bm{f}_i(t)$ and surface normal vector $\bm{n}_t(t)$ as $\alpha_i(t)$. To examine the temporal evolution of $\alpha_i(t)$, we initially consider a scenario involving only translation. Assuming the object undergoes a rigid translation $\bm{t}$ towards the equilibrium point, the initial force vector is given by:
\[\bm{f}_i(t_0) = k_i(\bm{o}_i - \bm{p}_i(t_0))\]
where $k_i$ is controller gain of fingertip. At any subsequent time, the force vector will be:
\[\bm{f}_i(t) = k_i(\bm{o}_i - \bm{p}_i(t_0) - \beta(t)\bm{t}) = \bm{f}_i(t_0) - k_i\beta(t)\bm{t}\] 
indicating that, during translation, the force vector changes from $\bm{f}_i(t_0)$ to $\bm{f}_i(t_0) - k_i\bm{t}$. The normal vector remains constant as translation does not affect the direction of the normal vector. Consequently, $\alpha_i(t)$ stays within the bounds defined by $\alpha_i(t_0)$ and $\alpha_i(t_\text{eq})$. Provided that the force directions at the initial and equilibrium states fall within the friction cone, the force direction at any transient state will also lie within this cone.

\subsubsection{Including rotation}
\begin{figure}[h]
  \centering
  \includegraphics[width=0.8\linewidth]{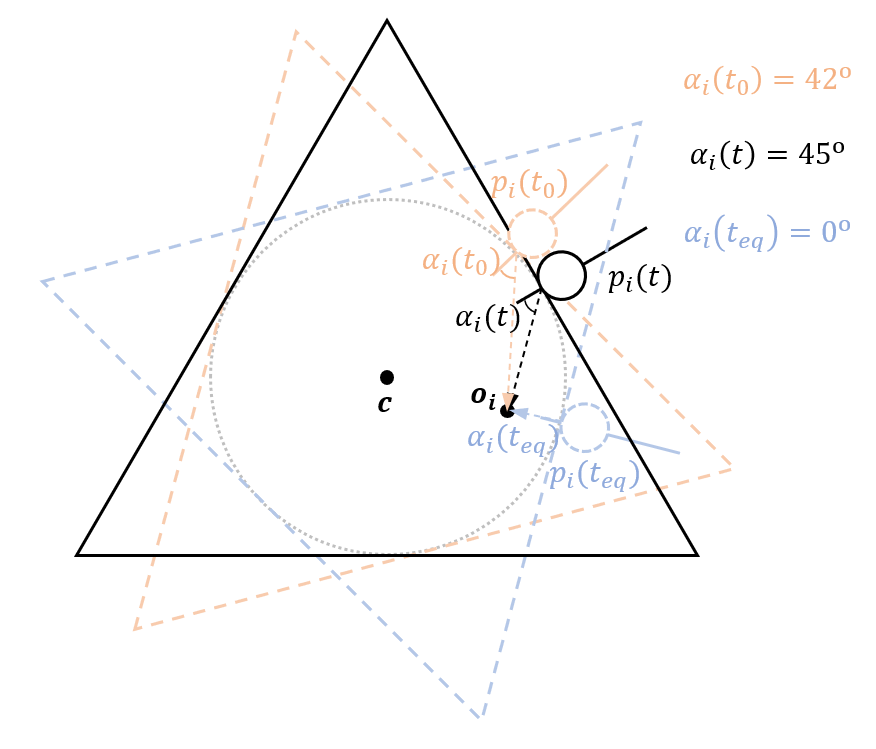}
  \caption{Triangle rotate around $\bm{c}$, at time $t$, the force vector is perpendicular to vector $\bm{c}-\bm{o}_i$ and $\alpha_i(t)=\frac{\pi}{4} = 45^o$. In initial state and equilibrium state: $\alpha_i(t_0)=42^o,\alpha_i(t_\text{eq})=0$}
  \label{fig:rotation}
\end{figure}

When the dynamic grasping process involves rotation, the feasibility of the initial and equilibrium state does not guarantee the feasibility of the transient states. Consider a triangle rotating around its center $\bm{c}$ (see Figure~\ref{fig:rotation}). We choose a target location $\bm{o}_i$, where the distance between $\bm{c}$ and $\bm{o}_i$ is $\frac{\sqrt{2}}{2}r$, with $r$ denoting the shortest distance from $\bm{c}$ to any edge of the triangle. Let the triangle rotate around its center until the force aligns with the surface normal vector. Then, $\alpha_i(t)$ initially increases to a maximum of $\alpha_\text{max} = \frac{\pi}{4}$ when $\bm{f}_i(t)$ is perpendicular to $\bm{o}_i - \bm{c}$, and then decreases to 0 at $t_{eq}$. If initially, angle $\angle \bm{p}_i(t_0)\bm{c}\bm{o}_i$  is less than $\frac{\pi}{4}$,$\alpha_i(t)$ will monotonically decrease and the force direction will always lies between the initial and equilibrium force directions. Bringing $\bm{o}_i$ closer to $\bm{c}$ reduces the change in $\alpha_i(t)$ for the same rotation angle on the object orientation $\theta(t)$, necessitating a greater rotation on $\theta(t)$ for $\alpha_i(t)$ to reach its maximum. Thus, adding $E_\text{tar}$ to the energy function Eq.\ref{eqn:energy} encourages the target location to stay deep inside the object, which allows more rotation on the object. Adding $E_\text{reg}$ to Eq. \ref{eqn:energy} regulates movement during the dynamic process and reduces the rotation during the dynamic process. In practice, the feasibility of transient states can usually be inferred from the feasibility of initial and equilibrium states.

\subsection{Collision spheres}
\label{sec:collisions}
\begin{figure}[h]
  \centering
   \includegraphics[width=0.8\linewidth]{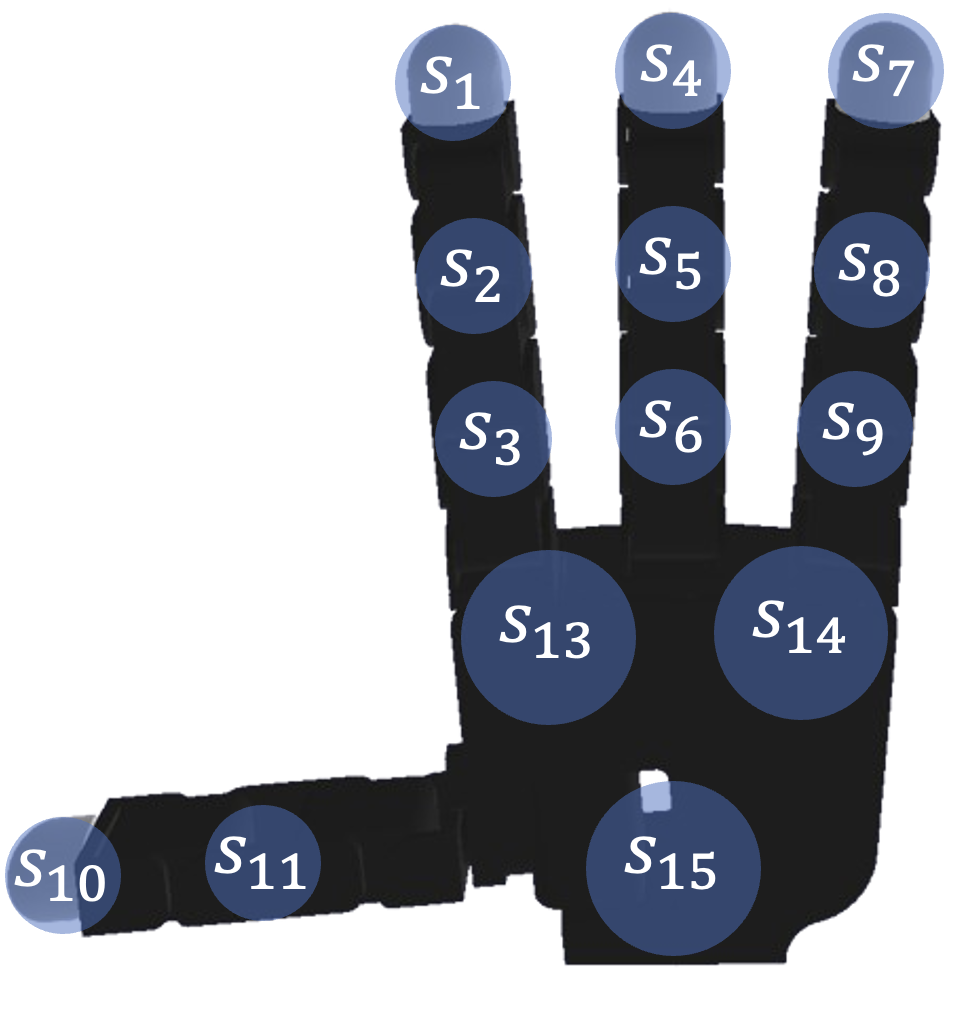}
  \caption{Placement of collision spheres for computing collision loss, each sphere has radius of 0.015}
  \label{fig:collision}
\end{figure}
Here we provide details on how we compute the energy terms $E_\text{self,ho,ht}$ in Eq~\ref{eqn:col} that are related to collisions with the hand itself, the object and table. Fig.~\ref{fig:collision} shows the placement of spheres that we use to approximate the geometry of the hand. $s_1\sim s_{11}$ have a radius of 1cm and $s_{13}\sim s_{15}$ have radius of 2cm. For computing self-collision energy $E_\text{self}$, we find it is sufficient to check the following collision pairs: $(s_1,s_4),(s_1,s_7),(s_1,s_{10}),(s_4,s_7),(s_4,s_{10}),(s_7,s_{10}),(s_2,s_5)$,\\$(s_5,s_8)$. When computing the hand object collision energy $E_\text{ho}$ and hand table collision energy $E_\text{ht}$, we use all collision spheres.

\subsection{Energy function of baseline method}
\label{sec:baseline}
For our baseline approach, we define the following energy function:
\[
    E= w_\text{fc}E_\text{fc} + w_\text{dist}E_\text{dist} + w_\text{col}E_\text{col} + w_\text{reg}E_\text{reg} + w_\text{uncer}E_\text{uncer}
\]\\
Most terms in this energy function are directly migrated from our method Eq.~\ref{eqn:energy}. We remove $E_\text{gain}$ as gains are choosen independently hence no other variables depends on the gains in the optimization problem. Therefore, the optimizer would drive the controller gains directly to the lowest possible value. We also remove $E_\text{tar}$ and $E_\text{force}$ as force and target locations are the result of the inner force closure solver of the baseline and therefore cannot be controlled by the outer loop. We set $w_\text{fc}=200$ and weights of other energy term is the same as Eq.~\ref{eqn:energy}. In \citeA{dex_graspnet, diff_wc}, $E_\text{fc}$ is approximated by assuming that each fingertip can only apply a force along the contact normal with a fixed magnitude, which accelerates computation. We replace it with a more accurate force closure metric as defined in \citeA{wu_bilevel} and compute it using the differentiable convex optimization solver cvxpylayers\citeA{cvxpylayers} as:
\[
    E_\text{fc} = \min{(||\sum_i^n \bm{f}_i||_2 + ||\sum_i^n \bm{p}_i\times\bm{f}_i||_2)}
\]
\[
    s.t.: \bm{f}_i\cdot\bm{n_i} \leq -\frac{1}{\sqrt{1+\mu^2}}||\bm{f}_i||, \bm{f}_i\cdot\bm{n_i} \leq -F_\text{min}\\
\]
As the grasps optimized by the baseline method do not involve a dynamic process,  we use $\bm{f}_i$, $\bm{p}_i$ without the time index. Because the force direction is determined by the force closure solver which cannot be controlled to reduce pregrasp uncertainty, we instead focus on reducing uncertainty at the fingertip contact location $\bm{p}_i$ and set the uncertainty energy term $E_\text{uncer}$ as follows: 
\[
    E_\text{uncer} = \sum_i d_\sigma(\bm{p}_i)
\]
Where $d_\sigma$ is variance value function of GPIS as shown in Sec.~\ref{subsec:gpis}.
\subsection{Grasp coverage experiment setup}
\label{sec:sim_exp}
\begin{figure}[h]
  \centering
  \includegraphics[width=\linewidth]{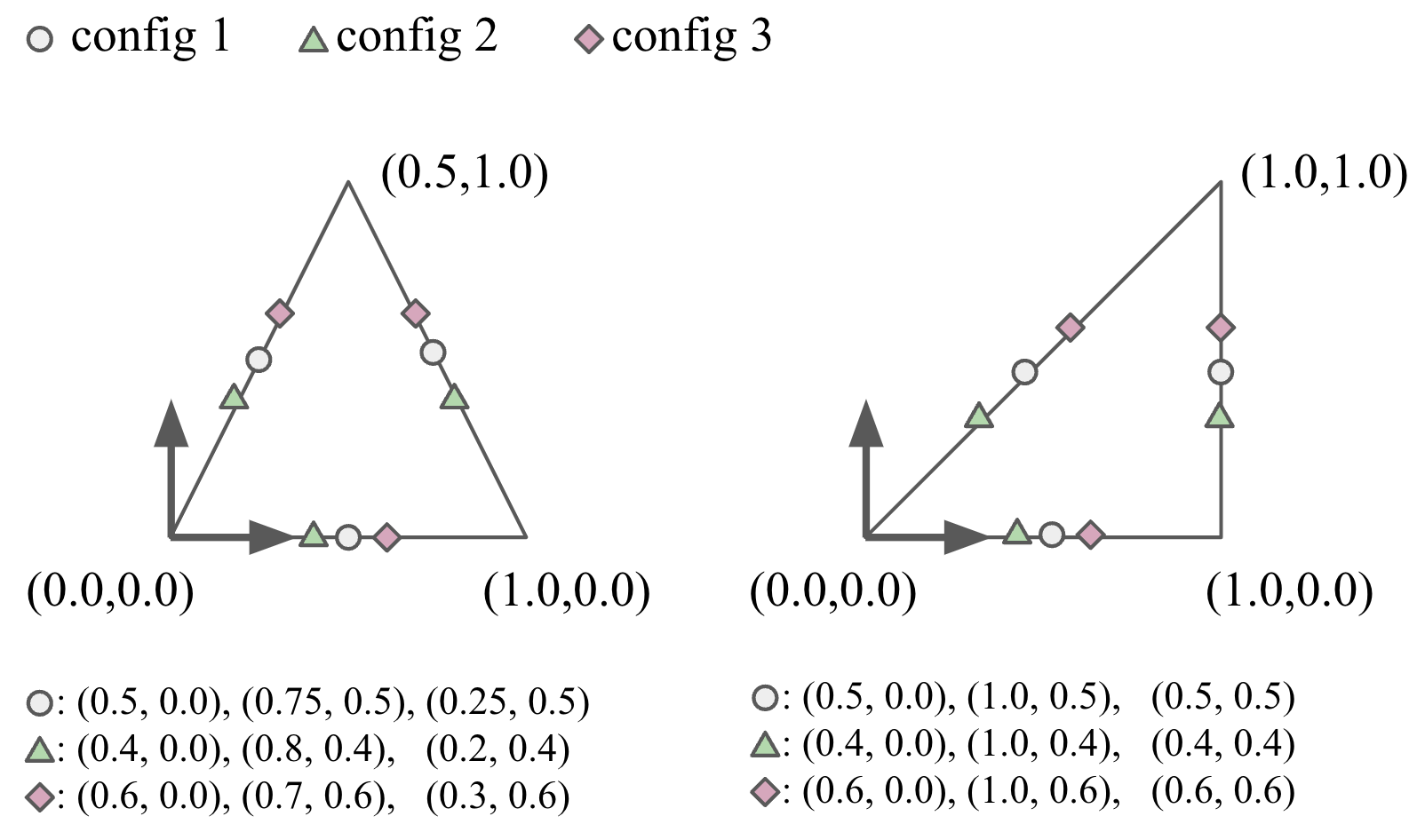}
  \caption{Specification of two triangles. Vertices are expressed in the local object frame. Contact points in each contact configurations are expressed as different markers.}
  \label{fig:triangles}
\end{figure}

\begin{table}[]
\begin{tabular}{l|ll}
\hline
             & Triangle 1                        & Triangle 2                      \\ \hline
config 1 & (0.5,0.0), (0.75,0.5), (0.25,0.5) & (0.5,0.0), (1.0,0.5), (0.5,0.5) \\
config 2 & (0.4,0.0), (0.8,0.4), (0.2,0.4)   & (0.4,0.0), (1.0,0.4), (0.4,0.4) \\
config 3 & (0.6,0.0), (0.7,0.6), (0.3,0.6)   & (0.6,0.0), (1.0,0.6), (0.6,0.6) \\ \hline
\end{tabular}
\caption{Summary of fingertip contact configurations}
\label{tab:tip_locations}
\end{table}
Fig. \ref{fig:triangles} shows dimension of two triangles and positions of different contact configurations we used in experiment \ref{subsec:coverage}. To ensure the best coverage of our method during optimization, we randomly initialize 2000 target positions and controller gains when optimizing grasp for each object pose.

\subsection{Clipping of spring grasp energy function}
\label{sec:clip_esp}
Here we illustrate implementation details of our spring grasp energy function $E_\text{eq}$. As the value of the contact margins $\epsilon_i(t_0),\epsilon_i(t_\text{eq})$ is in the range of $[-2,1]$, in rare cases if contact margins are below -1 and logarithm mapping becomes undefined, directly clipping the value of the margin will set the gradient of the energy function to zero. Inspired by Leaky Relu \citeA{leakyrelu}, we switch to an auxiliary energy to encourage the contact force to stay close to surface normal if contact margins are below -1. Therefore, $E_\text{sp}$ is defined as:
\[
\begin{split}
E_\text{sp} = -\sum_i^m
\begin{cases}\log(\epsilon_i(t_0)+1)& \epsilon_i(t_0)>-1\\
\log{\frac{\bm{f}_i(t_0)}{||\bm{f}_i(t_0)||^2_2}\cdot\bm{n}_i(t_0)}&\epsilon_i(t_0)\le-1
\end{cases}\\-\sum_i^m\begin{cases}\log(\epsilon_i(t_\text{eq})+1)&\epsilon_i(t_\text{eq})>-1\\\log{\frac{\bm{f}_i(t_\text{eq})}{||\bm{f}_i(t_\text{eq})||^2_2}\cdot\bm{n}_i(t_\text{eq})}&\epsilon_i(t_\text{eq})\le-1
\end{cases}
\end{split}
\]

\subsection{Fitting GPIS from partial point cloud}
We estimate the true surface of an object from a partial point cloud by fitting a GPIS to it. Following \citeA{most_similar}, we use three sets of points to fit GPIS: a) Surface points, b) Exterior points, and c) Interior points. We assign a different value and noise level to each set of points.
\subsubsection{Surface points}
We use points from the partial point cloud as surface points. As the object surface is represented by the zero-level set of GPIS, we assign each point in the point cloud the value 0. We set the noise level of each point to be 0.005m.
\subsubsection{Exterior points}
To generate exterior points, we initially determine the axis-aligned bounding box of the surface points and scale it by 120\% relative to the surface points' center. In total we have 14 exterior points which are located at each corner of the upscaled bounding box, as well as the midpoint of each bounding box edge. 14 points is sufficient for GPIS to distinguish between outside region and internal region. When fitting the GPIS, we found empirically that setting each exterior point value to be equal to half the length of the longest edge in the scaled bounding box works well. The noise level for each point is set to 0.2 meters.
\subsubsection{Interior points}
We compute interior points through convex combination of all surface points with random weights. To distribute these points evenly, rather than simply assigning weights from a uniform distribution to each surface point, we apply a softmax function to the initial random weights. This approach determines the actual weights for calculating interior points. For every object, we generate 50 interior points using this method. We found empirically that assigning each point a negative value equal to a quarter of the length of the longest edge of the upscaled bounding box works well. The noise level for each point is set at 0.05 meters.

\textcolor{black}{\subsection{Initial guesses for grasp planning}
We initialize 7 wrist poses around the center of the oriented bounding box of the observed partial point cloud. The 7 poses consist of 5 with the palm facing the table and 2 with the palm perpendicular to the table. Tab.\ref{tab:poses} illustrates different initial guesses. Moreover, we always initialize joint angles as the relaxed joint pose defined in \citeA{allegrokdl} and initialize the target position as a halfway point from each fingertip toward their center. Lastly, we use $k_1,k_2,k_3 = 80, k_4 = 160$ as initial controller gains.}

\begin{table}[t]
\centering
\resizebox{0.9\linewidth}{!}{
\begin{tabular}{r|l|l}
\hline
\multicolumn{1}{l|}{{\color{black} wrist pose}} & {\color{black} position(xyz)}      & {\color{black} orientation(Euler)} \\ \hline
{\color{black} 1}                               & {\color{black} -0.05, 0.0, 0.06}   & {\color{black} 0, 0, 0}            \\
{\color{black} 2}                               & {\color{black} -0.04, 0.03, 0.03}  & {\color{black} 0, 0, -45}          \\
{\color{black} 3}                               & {\color{black} -0.04, -0.03, 0.03} & {\color{black} 0, 0, 45}           \\
{\color{black} 4}                               & {\color{black} 0.1, 0.06, -0.05}   & {\color{black} -90, 90, 0}         \\
{\color{black} 5}                               & {\color{black} 0.0, 0.06,  0.05}   & {\color{black} -90, 0, 0}          \\
{\color{black} 6}                               & {\color{black} -0.0, -0.06, 0.03}  & {\color{black} 0, 0, 90}           \\
{\color{black} 7}                               & {\color{black} 0.02, -0.04, 0.03}  & {\color{black} 0, 0, 135}          \\ \hline
\end{tabular}}
\caption{\textcolor{black}{Initial wrist position and orientation, the unit of position is meter, and the unit of orientation is degree. We use intrinsic convention for Euler angles.}}
\label{tab:poses}
\end{table}

\textcolor{black}{\subsection{Partial successes}
To provide more information for Tab.\ref{tab:comparison}, we show the number of partial successes of each entry in Tab. \ref{tab:partialsuccess}.}

\begin{table*}
\centering
\resizebox{\linewidth}{!}{
\begin{tabular}{>{\color{black}}l|>{\color{black}}c>{\color{black}}c>{\color{black}}c|>{\color{black}}c|>{\color{black}}c|>{\color{black}}c>{\color{black}}c>{\color{black}}c}
\toprule
\multicolumn{1}{c|}{} & \thead{Ours \\ (3 views)}        & \thead{Ours \\ (2 views)}       &  \thead{Ours \\ (1 view)}       & \thead{w/o uncertainty \\ (1 view)} & \thead{w/o pregrasp \\ (1 view)} & \thead{Bilevel-high \\ (2 views)} & \thead{Bilevel-low \\ (2 views)} & \thead{Bilevel-heu \\ (2 views)} \\ \hline
Mustard bottle        & 1           & 0 & 0          & 3             & 1          & 2           & 1   &    1       \\
Lego                  & 1           & 1 & 0          & 1             & 0          & 1           & 3   &    1      \\
Pyramid               & 0           & 0 & 0          & 2             & 1          & 3           & 3   &    2        \\
Campell can           & 0           & 0 & 0          & 1             & 0          & 1           & 1   &    2         \\
Cheezit box           & 0           & 0 & 0          & 2             & 1          & 2           & 3   &    2     \\
Mug                   & 1           & 1 & 2          & 3             & 2          & 1           & 3   &    2       \\
Orange                & 0           & 2 & 1          & 0             & 0          & 1           & 2   &    0        \\
Coffee bottle         & 0           & 0 & 1          & 1             & 1          & 1           & 2   &    2      \\
Spam                  & 2           & 1 & 2          & 2             & 2          & 2           & 2   &    3      \\
Plane                 & 0           & 2 & 1          & 2             & 2          & 4           & 1   &    3      \\
Car                   & 1           & 1 & 1          & 3             & 3          & 2           & 2   &    4       \\
Banana                & 0           & 0 & 1          & 2             & 1          & 1           & 0   &    2      \\
Pear                  & 0           & 0 & 1          & 0             & 2          & 2           & 2   &    2       \\
Small box             & 0           & 0 & 0          & 1             & 3          & 2           & 2   &    1        \\ \hline
\textbf{Total}        & 6           & 8 & 10         & 23            & 19         & 25          & 27  &    27     \\ \bottomrule
\end{tabular}}
\caption{\textcolor{black}{Number of partial success for each experiment.}}
\label{tab:partialsuccess}
\end{table*}

\bibliographystyleA{plain}
\bibliographyA{ref_appd}
\end{document}